\documentclass[journal, 10pt, twocolumn]{IEEEtran}

\usepackage{graphicx} 
\usepackage{subfigure}
\usepackage{caption}
\usepackage{amsmath}
\usepackage{mathrsfs}
\usepackage{bm}
\usepackage{algorithm}
\usepackage{algorithmic}
\usepackage{comment}
\usepackage{multicol}
\usepackage{setspace}
\usepackage{pdfpages}
\usepackage{setspace}
\usepackage[numbers]{natbib}
\usepackage{multirow}
\usepackage{array}

\title{\LARGE {Training High-Performance and Large-Scale Deep Neural Networks with Full 8-bit Integers} 
}

\author{Yukuan Yang$^{a}$, Shuang Wu$^{a}$, Lei Deng$^{b*}$, Tianyi Yan$^{c}$, Yuan Xie$^{b}$, Guoqi Li$^{a*}$

\small $^a$Department of Precision Instrument, Center for Brain Inspired Computing Research, Tsinghua University, Beijing 100084, China. \\$^b$Department of Electrical and Computer Engineering, University of California, Santa Barbara, CA 93106, USA. \\$^c$School of Life Science, Beijing Institute of Technology, Beijing 100084, China.

\thanks{$^*$Corresponding authors. Email addresses: yyk17@mails.tsinghua.edu.cn (Y. Yang), wus15@mails.tsinghua.edu.cn (S. Wu), leideng@ucsb.edu (L. Deng), yantianyi@bit.edu.cn
 (T. Yan), Y. Xie (yuanxie@ece.ucsb.edu) and liguoqi@mail.tsinghua.edu.cn (G. Li).}
}
 
\begin{document}

\maketitle
\thispagestyle{empty}
\pagestyle{empty}

\begin{abstract}
Deep neural network (DNN) quantization converting floating-point (FP) data in the network to integers (INT) is an effective way to shrink the model size for memory saving and simplify the operations for compute acceleration. Recently, researches on DNN quantization develop from inference to training, laying a foundation for the online training on accelerators. However, existing schemes leaving batch normalization (BN) untouched during training are mostly incomplete quantization that still adopts high precision FP in some parts of the data paths. Currently, there is no solution that can use only low bit-width INT data during the whole training process of large-scale DNNs with acceptable accuracy. In this work, through decomposing all the computation steps in DNNs and fusing three special quantization functions to satisfy the different precision requirements, we propose a unified complete quantization framework termed as ``WAGEUBN'' to quantize DNNs involving all data paths including W (Weights), A (Activation), G (Gradient), E (Error), U (Update), and BN. Moreover, the Momentum optimizer is also quantized to realize a completely quantized framework. Experiments on ResNet18/34/50 models demonstrate that WAGEUBN can achieve competitive accuracy on the ImageNet dataset. For the first time, the study of quantization in large-scale DNNs is advanced to the full 8-bit INT level. In this way, all the operations in the training and inference can be bit-wise operations, pushing towards faster processing speed, decreased memory cost, and higher energy efficiency. Our throughout quantization framework has great potential for future efficient portable devices with online learning ability.
\end{abstract}

{ \it Keywords:}  Neural Network Quantization, 8-bit Training, Full Quantization, Online Learning Device


\section{INTRODUCTION}

Deep neural networks \cite{Lecun2015} have achieved state-of-art results in many fields like image processing \cite{Xie2012}, object detection \cite{Ren2015}, natural language processing \cite{Collobert2008}, and robotics \cite{zhang2016} through learning high-level features from a large amount of input data. However, due to the existence of a huge number of floating-point (FP) values and complex FP multiply-accumulate operations (MACs) in the process of network training and inference, the intensive memory overhead, large computational complexity, and high energy consumption impede the wide deployment of deep learning models. DNN quantization \cite{Sung2015} which converts FP MACs to bit-wise operations is an effective way to reduce the memory and computation costs and improve the speed of deep learning accelerators.

With the deepening of research, DNN quantization gradually transfers from inference quantization (BWN \cite{Courbariaux2015}, XNOR-Net \cite{Rastegari2016}, ADMM \cite{Cong2017}) to training quantization (DoReFa \cite{Zhou2016}, GXNOR-Net \cite{Deng2018}, FP8 \cite{Wang2018}). 
Usually, the inference quantization focuses on the forward pass; while the training quantization further quantizes the backward pass and weight updates. Recently, training quantization becomes a hot topic in the network compression community. Whereas, there are still two major issues in existing schemes. The first issue lies in the incomplete quantization, including two aspects: partial quantization and FP dependency. 
Partial quantization means that only parts of dataflows, not all of them, are quantized (e.g. DoReFa \cite{Zhou2016}, GXNOR-Net \cite{Deng2018} and QBP2 \cite{Banner2018}); FP dependency still remains FP values during the training process (e.g. MP \cite{Micikevicius2017} and FP8 \cite{Wang2018}). 
The second issue is that the quantization of batch normalization (BN) \cite{Ioffe2015} is ignored by most schemes (e.g. MP-INT \cite{Das2018} and FX Training \cite{Sakr2018}). BN is an essential layer for the training of DNNs by addressing the problem of the internal covariate shift of each layer's inputs, especially as the network deepens, allowing a much higher learning rate and less careful weight initialization.

Compared with all the studies above, WAGE \cite{Wu2018} is the most thorough work of DNNs quantization, which quantizes the data including W (Weights), A (Activation), G (Gradient), E (Error), U (Update) and replacing each BN layer with a constant scaling factor. WAGE has achieved competitive results on LeNet \cite{LeCun1998}, VGG \cite{Simonyan2014}, and AlexNet \cite{Krizhevsky2012}, providing a good inspiration for this work.  However, we find that WAGE is difficult to be applied in large-scale DNNs due to the absence of BN layers.  Besides, it is known that the gradient descent optimizer such as Momentum \cite{Qian1999} or Adam \cite{Kingma2014} increases the stability and even helps get rid of the local optimum, thus the speed and final performance are significantly improved. A complete quantization should cover the entire training process, including W, A, G, E, U, BN, and the optimizer. Regretfully, up to now, there is still no such solution that can achieve this complete quantization, especially on large-scale DNNs.

To address the issues of incomplete quantization and ignored BN quantization mentioned above and extend quantization framework to large-scale datasets and networks with high performance, we propose a unified complete quantization framework termed ``WAGEUBN" to constrain W, A, G, E, U, BN, and the optimizer in the low-bit integer (INT) space.
To the best of our knowledge, WAGEUBN is the first complete quantization framework achieving high performance in large-scale datasets, where all computation steps and operands in DNNs are decomposed and quantized.

We mainly make the following efforts to create the complete quantization framework. Firstly, according to the various data distributions and the role the quantized data plays in DNN training, we fuse three quantization functions to satisfy the different precision requirements. Furthermore, we propose a new storage and computing method by introducing a flag bit to expand the data coverage and solve the non-convergence problem caused by insufficient data representation. Last but not least, we quantize BN and Momentum optimizer for the first time, converting all FP operations in DNNs to bit-wise operations.
Compared with the full precision DNNs, DNNs under the full 8-bit WAGEUBN framework can achieve about $4\times$ memory saving. More importantly, the multiplication and accumulation operations of WAGEUBN, which are the main operations in DNNs, can perform $>$3$\times$ and 9$\times$ faster in speed, 10$\times$ and $>$30$\times$ lower in power, 9$\times$ and $>$30$\times$ smaller in circuit area, respectively. Besides, the efficient INT8 multiplication and accumulation operations also make WAGEUBN a big step ahead of most existing quantization schemes in computational costs, whether it is FP8\cite{Wang2018}, INT16\cite{Das2018}, FP16\cite{Micikevicius2017} or INT32. In addition to the huge advantages in memory cost, computing speed, energy consumption and circuit area, WAGEUBN also shows competitive accuracy on large-scale networks (ResNet18/34/50) and dataset (ImageNet \cite{Deng2009}). What's more, the hardware design of WAGEUBN is much simpler and more efficient because of the complete quantization. Due to the improvement of computing speed and saving in hardware resources and energy consumption, WAGEUBN provides a feasible idea for the architecture design of future efficient online learning chips used in portable devices with limited computational resources. The contributions of this work are twofold, which are summarized as follows:

\begin{itemize}
\item We address two main issues existing in most quantization schemes via fully quantizing all the data paths, including W, A, G, E, U, BN, and the optimizer, greatly reducing the memory and compute costs. What's more, we constrain the data to INT8 for the first time, pushing the training quantization to a new bit level compared with the existing FP16, INT16, and FP8 solutions.

\item Our quantization framework is validated in large-scale DNN models (ResNet18/34/50) over ImageNet dataset and achieves competitive accuracy with much fewer overheads, indicating great potential for future portable devices with online learning ability. 
\end{itemize}

The organization of this paper is as follows: Section \ref{Sec:RL} introduces the related work of DNN quantization; Section \ref{Sec:WAGEUBN} details the WAGEUBN framework; Section \ref{Sec:Exp} presents the experiment results of WAGEUBN and the corresponding analyses; Section \ref{Sec:Con} summarizes this work and delivers the conclusion.

\section{Related Work}
\label{Sec:RL}

With the wide applications of DNNs, the related compression technologies have been proposed rapidly, among which the quantization plays an important role. The development of DNN quantization can be divided into two stages, inference quantization and training quantization, according to the different quantization objects. 

\textbf{Inference quantization:} 
Inference quantization starts from constraining W into $\{-1,~1\}$ (BWN \cite{Courbariaux2015}), replacing complex FP MACs with simple accumulations. 
BNN \cite{Hubara2016} and XNOR-Net \cite{Rastegari2016} further quantize both W and A, making the inference computation dominated by bit-wise operations.
However, extremely low bit-width quantization usually leads to significant accuracy loss. For example, when the bit width comes to $<$4 bits, the accuracy degradation becomes obvious, especially for large-scale DNNs. Instead, the bit width of W and A for inference quantization can be reduced to 8 bits with little accuracy degradation. The study of inference quantization is sufficient for the deep learning inference accelerators. Whereas, this is not enough for efficient online learning accelerators because only the data in the forward pass are considered.

\textbf{Training quantization:} 
To further extend the quantization towards the training stage, DoReFa \cite{Zhou2016} trains DNNs with low bit-width W, A, and G, while leaving E and BN unprocessed. MP \cite{Micikevicius2017} and MP-INT \cite{Das2018} use FP16 and INT16 values, respectively, to constrain W, A, and G. Recently, FP8 \cite{Wang2018} further pushes W, A, G, E, and U to 8, 8, 8, 8, and 16-bit FP values, respectively, still leaving BN untouched. 
QBP2 \cite{Banner2018} replaces the conventional BN with range BN and constrains W, A, and E to INT8 values while calculating G with FP MACS.
Recently, WAGE \cite{Wu2018} adopts a layer-wise scaling factor instead of using the BN layer and quantizes W, A, G, E, and U to 2, 8, 8, 8, and 8 bits, respectively. Despite its thorough quantization, WAGE is difficult to be applied to large-scale DNNs due to the absence of powerful BN layers. 
In summary, there still lacks a complete INT8 quantization framework for the training of large-scale DNNs with high accuracy.

\section{WAGEUBN Framework}
\label{Sec:WAGEUBN}

The main idea of WAGEUBN is to quantize all the data in DNN training to INT8 values. In this section, we detail the WAGEUBN framework implemented in large-scale DNN models. The organization of this section is as follows: Subsection \ref{STE} introduces the straight-through estimator (STE) \cite{hinton2012neural,bengio2013estimating} method which is accepted and used by most researchers to solve the non-differentiable problem of quantization; Subsection \ref{Notations} and Subsection \ref{Quantization functions} describe the notations and quantization functions, respectively; Subsection \ref{Quantization details of WAGEUBN} explains the specific quantization schemes for W, A, G, E, U, BN, and the Momentum optimizer, respectively; Subsection \ref{Quantization Framework} goes through the overall implementation of WAGEUBN, including in both forward and backward passes; Subsection \ref{Process of Algorithm} summarizes the whole process and shows the pseudo codes.

\subsection{Straight-through estimator}
\label{STE}

In the early stage of the study, the non-differentiable problem in mathematical sense caused by quantization has been hindering the development of quantization research. However, since the straight-through estimator (STE) method was used to estimate the gradient of quantized data in BNN \cite{Hubara2016}, almost all the works in the field of quantization have adopted this method to avoid the mathematical non-derivative problem \cite{Zhou2016,hubara2017quantized,zhou2017balanced,he2016effective}.

The STE method used in WAGEUBN can be illustrated as the following 
\begin{eqnarray}
\begin{split}
Forward: \bm{x_q} = Q(\bm{x}) \\
Backward: \dfrac{\partial L}{\partial \bm{x}} = \dfrac{\partial L}{\partial \bm{x_q}}
\label{ste}
\end{split}
\end{eqnarray}
where $\bm{x}$ is the data to be quantized, $Q(\cdot)$ is the quantization function that may be non-differentiable, and $L$ denotes the objective function.

\subsection{Notations}
\label{Notations}

\begin{figure}[!htpb]
    \centering
    \includegraphics[width=0.5\textwidth]{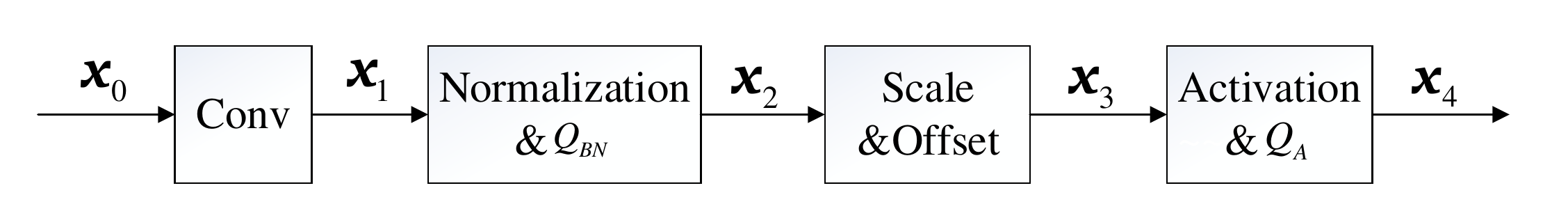}
    \caption {Forward quantization of the $l$-th layer in DNNs.}
    \label{fig:forward}
\end{figure}

Before introducing the WAGEUBN quantization framework formally, we need to define some notations. Considering the $l$-th layer of DNNs, we divide the forward pass of DNNs into four steps as described in Figure \ref{fig:forward} (BN is divided into two steps: Normalization \& $Q_{BN}$ and Scale \& Offset).

Specifically, we have
\begin{eqnarray}
\centering
\begin{split}
\rm{Input:~~~~~~~~~}~~ &\bm{x}_{0}^{l} = \bm{x}_{4}^{l-1}\\
\rm{Conv:~~~~~~~~~}~~ &\bm{x}_{1}^{l} = \bm{W}_{q}^{l} \bm{x}_{0}^{l}\\
\rm{Normalization \& Q_{BN}:}~~ &\bm{x}_{2}^{l} =Q_{BN}(\dfrac{\bm{x}_{1}^{l}-\mu^{l}_{q}}{\sigma^{l}_{q}})\\
 \rm{Scale\&Offset:~~~~}~~ &\bm{x}_{3}^{l} = \bm{\gamma}_{q}^{l}\bm{x}_{2}^{l} + \bm{\beta}_{q}^{l}\\
\rm{Activation\&Q_{A}:~~}~~ &\bm{x}_{4}^{l} = Q_{A}\left[ relu(\bm{x}_{3}^{l}) \right ]
\end{split}~~~,
\label{X_define}
\end{eqnarray}
where $\bm{x}^{l-1}_{4}$ is the output of the $(l-1)$-th layer and $\bm{x}^{l}_{0}$ is the input of the $l$-th layer; 
$\bm{W}_{q}^{l}$ is the quantized weight for convolution;  $Q_{BN}$ is the quantization function used for constraining $\bm{x}_{2}^{l}$ to low-bit INT values, which will be given in Equation (\ref{q_define}); $\mu^{l}_{q}$ and $\sigma^{l}_{q}$ are the quantized mean and standard deviation value of one mini-batch;  $\bm{\gamma}_{q}^{l}$ and $\bm{\beta}_{q}^{l}$ are the quantized scale and bias used in the BN layer of DNNs;  $Q_{A}$ is the quantization function for activation which will be detailed in Equation (\ref{q_A});  $relu$ is the activation function which is commonly used in DNNs. Noting that every step in the forward pass is quantized, $\bm{x}_{0}^{l}, \bm{x}_{1}^{l}, \bm{x}_{2}^{l}, \bm{x}_{3}^{l}, \bm{x}_{4}^{l}$ are all integers.  In order to make the notations used in the paper consistent, we make the following rules: subscript $q$ denotes the data which have been quantized to INT values and superscript $l$ denote the layer index.

\begin{figure}[htpb]
    \centering
    \includegraphics[width=0.5\textwidth]{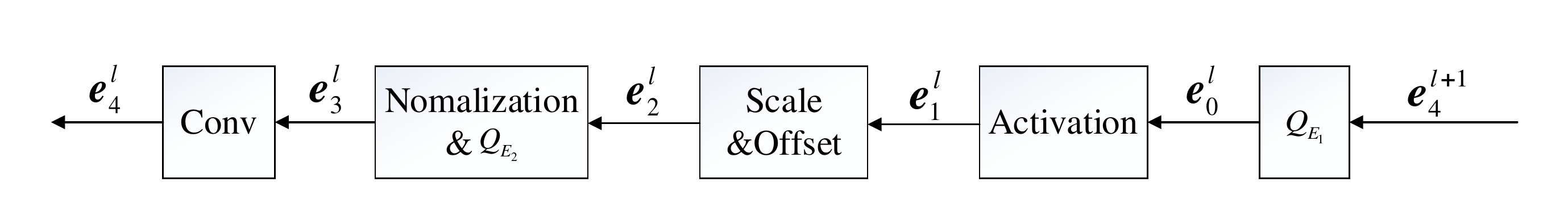}
    \caption {Backward quantization of the $l$-th layer in DNNs.}
    \label{fig:backward}
\end{figure}

Different from most existing schemes, we define $\bm{e}$ and $\bm{g}$ respectively, where $\bm{e}$ represents the gradient of A (activation) which is used in the error backpropagation and $\bm{g}$ represents the gradient of W (weights) which is used in the weight update. Moreover, we quantize the BN layers, including both the forward and backward passes, which is not well touched in most prior work.  Similar to the forward pass, we divide the backward pass of the $l$-th layer into five steps as shown in Figure \ref{fig:backward}. According to the derivative chain rules, we have
\small
\begin{eqnarray}
\centering
\begin{split}
\rm{Q_{E_{1}}:~~~~~~~~~~}&\bm{e}_{0}^{l} = Q_{E_{1}}(\dfrac{\partial L}{\partial \bm{x}_{4}^{l}}) = Q_{E_{1}}(\bm{e}_{4}^{l+1})\\
\rm{Activation:~~~~~~~}&\bm{e}_{1}^{l} = \dfrac{\partial L}{\partial \bm{x}_{3}^{l}} = \dfrac{\partial L}{\partial \bm{x}_{4}^{l}} \odot \dfrac{\partial \bm{x}_{4}^{l}}{\partial \bm{x}_{3}^{l}} = \bm{e}_{0}^{l} \odot \dfrac{\partial \bm{x}_{4}^{l}}{\partial \bm{x}_{3}^{l}}\\
\rm{Scale\&Offset:~~~~~}&\bm{e}_{2}^{l} = \dfrac{\partial L}{\partial \bm{x}_{2}^{l}}=\dfrac{\partial L}{\partial \bm{x}_{3}^{l}} \odot \dfrac{\partial \bm{x}_{3}^{l}}{\partial \bm{x}_{2}^{l}} = \bm{e}_{1}^{l} \odot \gamma_{q}^{l}\\
\rm{Norm \& Q_{E_{2}}:~}&\bm{e}_{3}^{l} = Q_{E_{2}}(\dfrac{\partial L}{\partial \bm{x}_{1}^{l}}) =Q_{E_{2}}(\dfrac{\partial L}{\partial \bm{x}_{2}^{l}} \odot \dfrac{\partial \bm{x}_{2}^{l}}{\partial \bm{x}_{1}^{l}})\\
 &~~~= Q_{E_{2}}(\bm{e}_{2}^{l} \odot \dfrac{\partial \bm{x}_{2}^{l}}{\partial \bm{x}_{1}^{l}})\\
\rm{Conv:~~~~~~~~~}&\bm{e}_{4}^{l} = \dfrac{\partial L}{\partial \bm{x}_{0}^{l}} =  \dfrac{\partial \bm{x}_{1}^{l}}{\partial \bm{x}_{0}^{l}}  \dfrac{\partial L}{\partial \bm{x}_{1}^{l}} = {W_{q}^{l}}^{T} \bm{e}_{3}^{l}
\label{E_define}
\end{split} 
\end{eqnarray}
\normalsize
where $L$ is the loss function, $\bm{e}_{4}^{l+1}$ represents the error from the $(l+1)$-th layer, and $\odot$ represents the Hadamard product.  For vectors with the same dimension, such as $\bm{a}=(a_{1},a_{2},\cdots, a_{n})$ and $\bm{b}=(b_{1},b_{2},\cdots, b_{n})$, we have: $\bm{a} \odot \bm{b}=(a_{1}b_{1},a_{2}b_{2},\cdots, a_{n}b_{n})$. Two quantization functions are used here: $Q_{E_{1}}$ is the quantization function detailed as Equation (\ref{16bitE}) that converts high bit-width integers to low bit-width integers; $Q_{E_{2}}$ detailed as Equation (\ref{8bitE2}) is trying to convert FP values to low bit-width integers. ${W_{q}^{l}}^{T}$ is the transposed matrix of $W_{q}^{l}$, and $\partial \bm{x}_{4}^{l} / \partial \bm{x}_{3}^{l}$ represents the gradient of activation. When $relu$ is used as the activation function, $\partial \bm{x}_{4}^{l} / \partial \bm{x}_{3}^{l}$ is a tensor containing only 0 and 1 elements. 

According to the definitions given above, the gradients of W, $\bm{\gamma}$, and $\bm{\beta}$ can be summarized as follows
\small
\begin{eqnarray}
\centering
\begin{split}
\bm{g}^{l}_{W} &= \dfrac{\partial L}{\partial {W}^{l}} = \dfrac{\partial L}{\partial \bm{x}_{1}^{l}} \dfrac{\partial \bm{x}_{1}^{l}}{\partial {W}^{l}} = \bm{e}_{3}^{l}{\bm{x}_{0}^{l}}^{T} \\
\bm{g}^{l}_{\bm{\gamma}} &= \dfrac{\partial L}{\partial \bm{\gamma}^{l}} = \dfrac{\partial L}{\partial \bm{x}_{3}^{l}} \odot \dfrac{\partial \bm{x}_{3}^{l}}{\partial \bm{\gamma}^{l}} = \bm{e}_{1}^{l} \odot \bm{x}_{2}^{l} \\
\bm{g}^{l}_{\bm{\beta}} &= \dfrac{\partial L}{\partial \bm{\beta}^{l}} = \dfrac{\partial L}{\partial \bm{x}_{3}^{l}} \odot \dfrac{\partial \bm{x}_{3}^{l}}{\partial \bm{\beta}^{l}} = \bm{e}_{1}^{l}. \\
\end{split} 
\label{G_define}
\end{eqnarray}
\normalsize

To further reduce the bit width of G that will increase greatly after the multiplication, we have
\begin{eqnarray}
\centering
\begin{split}
&\bm{g}^{l}_{Wq} = Q_{G_{W}}(g^{l}_{W}) \\
&~\bm{g}^{l}_{\bm{\gamma}q} = Q_{G_{\bm{\gamma}}}(g^{l}_{\bm{\gamma}}) \\
&~\bm{g}^{l}_{\bm{\beta}q} = Q_{G_{\bm{\beta}}}(g^{l}_{\bm{\beta}}) \\
\end{split} 
\label{GQ_define}
\end{eqnarray}
where $Q_{G_{W}}$, $Q_{G_{\bm{\gamma}}}$, and $Q_{G_{\bm{\beta}}}$ are quantization functions for the gradient of W, ${\bm{\gamma}}$, and ${\bm{\beta}}$, respectively, which will be shown in Equation (\ref{QG_define}).

Some notations to be used below are also explained here. $k_{W}$, $k_{A}$, $k_{G_{W}}$, $k_{E}$ ($k_{E_{1}}$ and $k_{E_{2}}$), and $k_{BN}$ are the bit width of W, A, G, E, and BN, respectively. $k_{WU}$, $k_{\bm{\gamma}U}$, and $k_{\bm{\beta}U}$ are the bit width of W, $\bm{\gamma}$, and $\bm{\beta}$ update, which are also the bit width of data stored in memory. $k_{\bm{\gamma}}$, $k_{\bm{\beta}}$, $k_{\mu}$, and $k_{\sigma}$ are the bit width of $\bm{\gamma}$, $\bm{\beta}$, $\mu$, and $\sigma$, respectively, used in the BN layer. $k_{G\bm{\gamma}}$ and $k_{G\bm{\beta}}$ are the bit width of $\bm{\gamma}$ and $\bm{\beta}$ gradient, respectively. $k_{Mom}$ and $k_{Acc}$ are the bit width of momentum coefficient ($Mom$) and accumulation ($Acc$), respectively, used in the Momentum optimizer. At last, $k_{lr}$ is the bit width of the learning rate.

\subsection{Quantization Functions}
\label{Quantization functions}

There are three quantization functions used in WAGEUBN. The direct-quantization function uses the nearest fixed-point values to represent the continuous values of W, A, and BN. The constant-quantization function for G is used to keep the bit width of U (update) fixed since G is directly related to U. Because U and the weights stored in memory have the same bit width, the bit width of weights stored in memory can be fixed, which is more hardware-friendly. The magnitude of E is very small, so the shift-quantization function reduces the bit width of E greatly compared with the direct-quantization function under the same precision.

~\\
\indent \textit{(1) Direct-quantization function}
\vspace{4pt}

The direct-quantization function simply approximates a continuous value to its nearest discrete state and is defined as
\begin{equation}
Q(\bm{x}, k) = \dfrac{round(\bm{x} \cdot 2^{k-1}) }{2^{k-1}}
\label{dqf}
\end{equation}
where $k$ is the bit width, and $round(\cdot)$ rounds a number to its nearest INT value.

~\\
\indent \textit{(2) Constant-quantization function}
\vspace{4pt}

The intention of constant-quantization function is to normalize a tensor firstly, then limit it to INT, and finally maintain its magnitude. 
It is governed by
\begin{eqnarray}
\begin{split}
&~~~~~~~~~~~~~~~R(\bm{x}) = 2^{round(\log_2 ({max(|\bm{x}|))})}\\
&~~~~~~~~~~~~~~Sr(x)= \left\{
\begin{array}{lr}
\lfloor x \rfloor, ~~~   P_{x} = \lceil x \rceil -x   \\
\\
\lceil x \rceil, ~~~ P_{x} = x- \lfloor x \rfloor \\ 
\end{array} 
\right. \\
&~~~~~~~~~~~~~~~~~~~~Norm(\bm{x})=\dfrac{\bm{x}}{R(\bm{x})}\\
&Sd(\bm{x}) = clip\left\{ Sr\left[ dr \cdot Norm(\bm{x}) \right], -dr+1, dr-1 \right\} \\ 
&~~~~~~~~~~~~~~~~~~~~CQ(\bm{x}) = \dfrac{Sd(\bm{x})}{2^{k_{GC}-1}}.\\
\end{split}
\label{cqf}
\end{eqnarray}

\begin{figure*}[!htpb]
    \centering
    \includegraphics[width=0.99\textwidth]{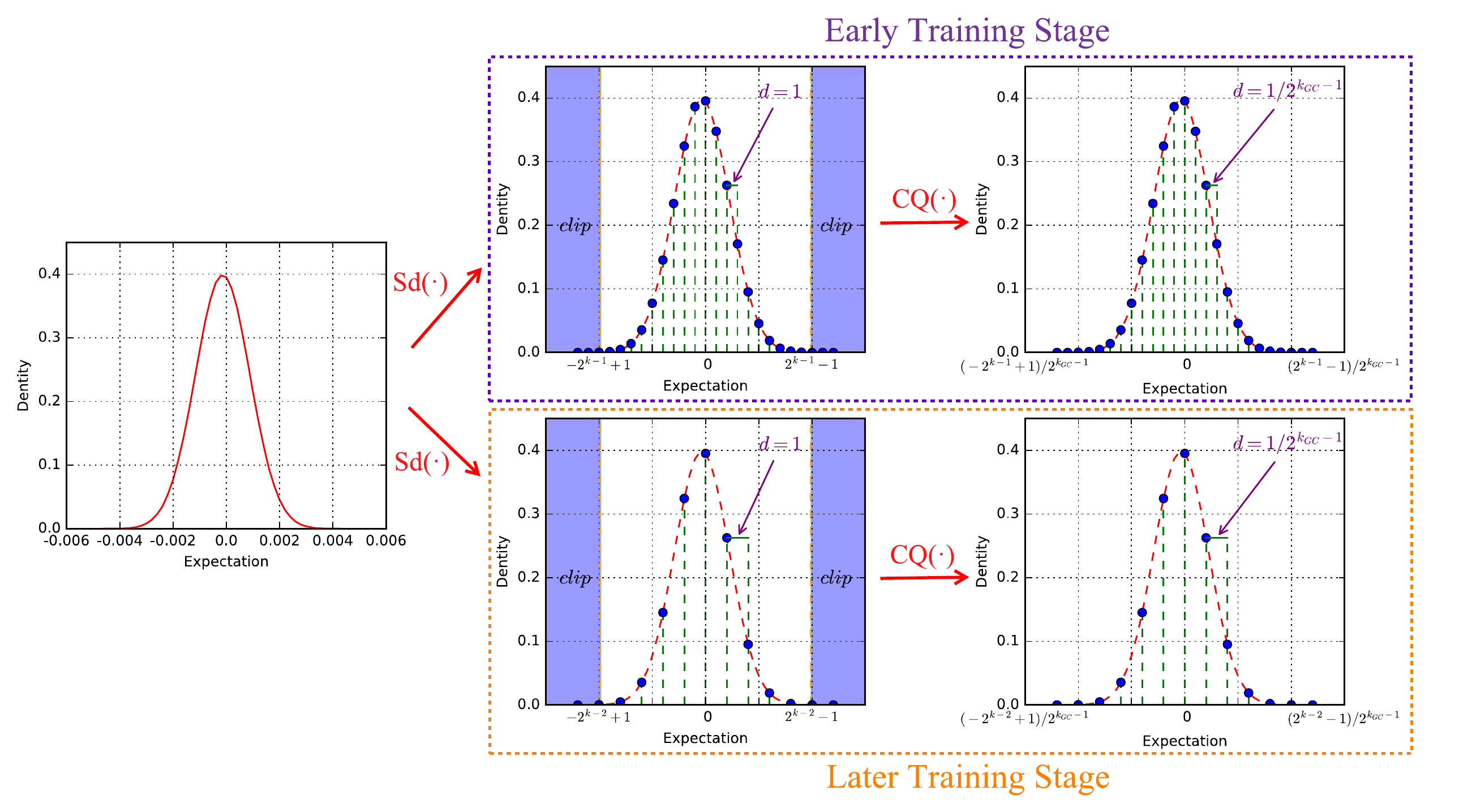}
    \caption {Illustration of the constant-quantization function during training.}
    \label{fig:CQ}
\end{figure*}

The illustration of constant-quantization function is described in Figure \ref{fig:CQ}. Here, $R(\cdot)$ is used to project the maximum value of $\bm{x}$ to its nearest fixed-point value, which is prepared for normalization; $Sr(\cdot)$ is a stochastic rounding function used for converting a continuous float value to its nearby INT value in a probabilistic manner and $P_{x}$ is the rounding probability; $Norm(\cdot)$ denotes normalization and $clip(\cdot)$ is a saturation function limiting the data range;$Sd(\cdot)$ is to shift the distribution of $\bm{x}$ and limit $\bm{x}$ to INT values between $-dr+1$ and $dr-1$. Here $dr \in [2^{k-1}, 2^{k-2}, ..., 1]$ limits the data range after mapping and decreases as the training goes on, presenting the same effect as reducing the learning rate. For example, $Sd(\cdot)$ maps G to \{-127, -126, $\cdots$, 126, 127\} and \{-63, -62, $\cdots$, 62, 63\} in the early training stage ($k=8$, $dr=128$, epoch in $[0,30]$) and later training stage ($k=7$, $dr=64$, epoch in $[30,60]$), respectively. $CQ(\cdot)$ is utilized to maintain the magnitude order of data, where $2^{k_{GC}-1}$ is a constant scaling factor and $k_{GC}$ is its bit width.

~\\
\indent \textit{(3) Shift-quantization function}
\vspace{4pt}

The shift-quantization function serves for the quantization of E and is defined as
\small
\begin{equation}
\arraycolsep=1pt\def\arraystretch{1.0}
\begin{array}{rl}
&~~~~~~~~~~~~~~~~~~~~~~~~~~~~~~ d(k) = \dfrac{1}{2^{k-1}} \\\\
&SQ(\bm{x}, k) = R(\bm{x}) \cdot clip \left\{ Q \left [Norm(\bm{x}), k \right], -1+d(k), 1-d(k)\right\}
\label{sqf}
\end{array}
\end{equation}

\normalsize
\noindent where $d(\cdot)$ is the minimum interval for a k-bit INT and $Q(\cdot)$ is the direct-quantization function defined in Equation (\ref{dqf}). 

The shift-quantization function normalizes E first, then converts E to fixed-point values, and finally uses a layer-wise scaling factor ($R(\cdot)$ defined in Equation (\ref{cqf})) to maintain the magnitude.  The differences between the constant-quantization function and the shift-quantization function mainly exist in two points: First, the constant-quantization uses a constant to keep the magnitude for hardware friendliness while the shift-quantization uses a lay-wise scaling factor; Second, the constant-quantization contains a stochastic rounding process while the shift-quantization function does not.

\subsection{Quantization Schemes in WAGEUBN}
\label{Quantization details of WAGEUBN}

After introducing the quantization functions used in our WAGEUBN framework, we provide detailed quantization schemes.

~\\
\textit{(1) Weight Quantization} 
\vspace{4pt}

\indent Since weights are stored and used as fixed-point values, weights should be also initialized discretely. An initialization method proposed by MSRA \cite{He2015} has been evidenced helpful for faster training. The initialization of weights can be formulated as follows
\small
\begin{eqnarray}
\begin{split}
&~~~~~~~~~~~~~~~~~~~W^{'}\sim N(0, \dfrac{1}{\sqrt{n_{in}}})\\
&W=clip \left[ Q(W^{'}, k_{WU}), -1+d(k_{WU}), 1-d(k_{WU}) \right]
\end{split}
\label{w_init}
\end{eqnarray}
\normalsize
where $n_{in}$ is the layer's fan-in number, and $k_{WU}$ is the bit width of weight update and the memory storage. 

Because of the different bit width for weight storage and computation, it should be quantized from $k_{WU}$ bits to $k_{W}$ (the bit width of weights used for convolution) bits for convolution. In addition, we also limit the data range of W. Finally, the quantization function for W is
\small
\begin{equation}
Q_{W}(\bm{x})=clip\left[ Q(\bm{x}, k_{W}), -1+d(k_{W}), 1-d(k_{W}) \right ].
\label{W_Q}
\end{equation}
\normalsize

\indent \textit{(2) Batch Normalization Quantization} 
\vspace{4pt}

\indent As aforementioned, BN plays an important role in training large-scale DNNs. WAGE \cite{Wu2018} has proved that simple scaling layers are not enough to replace BN layers. Conventional BN layer can be divided into two steps as
\begin{eqnarray}
\begin{split}
\hat{\bm{x}}=&\dfrac{\bm{x}-\mu^{l}}{\sqrt{{\sigma^{l}}^2+\epsilon}}\\
\bm{y}=&\bm{\gamma}^{l}\hat{\bm{x}}+\bm{\beta}^{l}
\end{split}
\label{c_BN}
\end{eqnarray}
where $\mu^{l}$ and $\sigma^{l}$ are the mean and standard, respectively, deviation of $\bm{x}$ over one mini-batch in the $l$-th layer; $\epsilon$ is a small positive value added to $\sigma$ to avoid the case of dividing by zero; $\bm{\gamma^{l}}$ and $\bm{\beta^{l}}$ are the scale and offset parameters, respectively.

Under the WAGEUBN framework, the BN layer is also quantized. Through the operations described in Equation (\ref{q_BN}), all operands are quantized and all operations are bit-wise. Specifically, the quantization follows
\begin{eqnarray}
\begin{split}
\mu_{q}^{l}=Q_{\mu}(\mu^{l}) , ~~ & \sigma_{q}^{l}=Q_{\sigma}(\sigma^{l}) \\
\hat{\bm{x}}=Q_{BN}&(\dfrac{\bm{x}-\mu_{q}^{l}}{\sigma_{q}^{l}+\epsilon_{q}})\\
\bm{\gamma}_{q}^{l}=Q_{\bm{\gamma}}(\bm{\gamma}^{l}) , ~~ & \bm{\beta}_{q}^{l}=Q_{\bm{\beta}}(\bm{\beta}^{l}) \\
\bm{y}=\bm{\gamma}^{l}_{q}&\hat{\bm{x}}+\bm{\beta}^{l}_{q}
\end{split}
\label{q_BN}
\end{eqnarray}
where $Q_{\mu}, Q_{\sigma}, Q_{\bm{\gamma}}, Q_{\bm{\beta}}, Q_{BN}$ are the quantization functions converting the operands to fixed-point values defined as
\begin{eqnarray}
\begin{split}
Q_{\mu}(\bm{x}) = &Q(\bm{x}, k_{\mu}), Q_{\sigma}(\bm{x}) =Q(\bm{x}, k_{\sigma}) \\
Q_{\bm{\gamma}}(\bm{x})= &Q(\bm{x}, k_{\bm{\gamma}}), Q_{\bm{\beta}}(\bm{x})=Q(\bm{x}, k_{\bm{\beta}}) \\
&Q_{BN}(\bm{x}) =Q(\bm{x}, k_{BN}).
\end{split}
\label{q_define}
\end{eqnarray}
And $\epsilon_{q}$ is a small fixed-point value, playing the same role as $\epsilon$ in Equation (\ref{c_BN}); $k_{\mu}, k_{\sigma}, k_{\bm{\gamma}}, k_{\bm{\beta}}, k_{BN} $ are the bit width of $\mu, \sigma, \bm{\gamma}, \bm{\beta}$ and $\hat{\bm{x}}$, respectively.

~\\
\indent \textit{(3) Activation Quantization}
\vspace{4pt}

After the convolution and BN layers in the forward pass, the bit width of operands increases due to the multiplication operation. To reduce the bit width and keep the input bit width of each layer consistent, we need to quantize the activations. Here, the quantization function for activations can be described as
\begin{equation}
Q_{A}(\bm{x})=Q(\bm{x}, k_{A})
\label{q_A}
\end{equation}
where $k_{A}$ is the bit width of activations.

~\\
\indent \textit{(4) Error Quantization}
\vspace{4pt}

In Equation (\ref{E_define}), we have given the definition of E and quantized E. Through investigating the importance of error propagation in DNN training, we find that the quantization of E is very essential for the model convergence. If E is naively quantized using the direct-quantization function, it will require a large bit width of operands to realize the convergence of DNNs. Instead, we use the following shift-quantization function
\begin{eqnarray}
\begin{split}
Q_{E_{1}}(\bm{x})=SQ(\bm{x}, k_{E_{1}}) \\
\end{split}
\label{16bitE}
\end{eqnarray}
where $SQ(\cdot)$ is the shift-quantization function defined in Equation (\ref{sqf}), and $k_{E_{1}}$ is the bit width of $\bm{e}_{0}^{l}$ defined in Equation (\ref{E_define}).

As mentioned above, we use $Q_{E_{1}}$ and $Q_{E_{2}}$ for the error quantization. However, the precision requirements of $Q_{E_{1}}$ and $Q_{E_{2}}$ vary a lot. Experiments show that $k_{E_1}=8$ affects little on accuracy while $k_{E_{2}}\leq8$ will cause the non-convergence of large-scale DNNs when using $SQ(\cdot)$ as the quantization function. $k_{E_{2}}=16$ is a proper value for the training of DNNs with minimum accuracy degradation. More analyses will be given in Subsection \ref{BWE}. Here we will provide two versions of $Q_{E_{2}}$, the 16-bit and 8-bit versions. The 16-bit $Q_{E_{2}}$ is defined as
\begin{equation}
Q_{E_{2}}(\bm{x})=SQ(\bm{x}, k_{E_{2}}) \\
\label{16bitE2}
\end{equation}
where $k_{E_{2}}$ is the bit width of $\bm{e}_{3}^{l}$ defined in Equation (\ref{E_define}).

Experiments have proved the data range covered by 8-bit $Q_{E_{2}}$ ($k_{E_{2}}=8$)  is not sufficient to train DNNs. In order to expand the coverage of quantization function while still maintaining a low bit width,  we introduce a layer-wise scaling factor $Sc$ and a flag bit. Then, to distinguish it from $Q_{E_2}$ defined in Equation (\ref{16bitE2}), we name the quantization function Flag $Q_{E_2}$ and the quantization process is governed by
\small
\begin{eqnarray}
\begin{split}
&~~~~~~~~~~~~~~~~~~~~~~~~~~~~Sc = \dfrac{R(\bm{x}) }{2^{k_{E_2}-1}} \\
&Q_{E_2}(\bm{x})=\left\{
\begin{array}{lr}
 Sc\cdot clip\left[round(\dfrac{\bm{x}}{Sc}), min, max\right], | \dfrac{\bm{x}}{Sc} | \geq 1 \\
\\
Sc \cdot Q(\dfrac{\bm{x}}{Sc}, k_{E_{2}}) , | \dfrac{\bm{x}}{Sc}| < 1 \\ 
\end{array}
\right.
\end{split}
\label{8bitE2}
\end{eqnarray}
\normalsize
where $k_{E_2}=8$, $min=-2^{k_{E_2}}+1$, and $max= 2^{k_{E_2}}-1$.

\begin{figure}[htpb]
    \centering
    \subfigure[]{\includegraphics[width=0.48\textwidth]{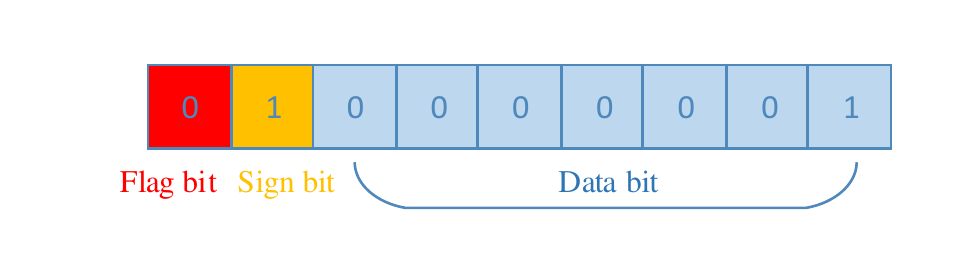}
        \label{fig:2a}}
    \subfigure[]{
        \includegraphics[width=0.48\textwidth]{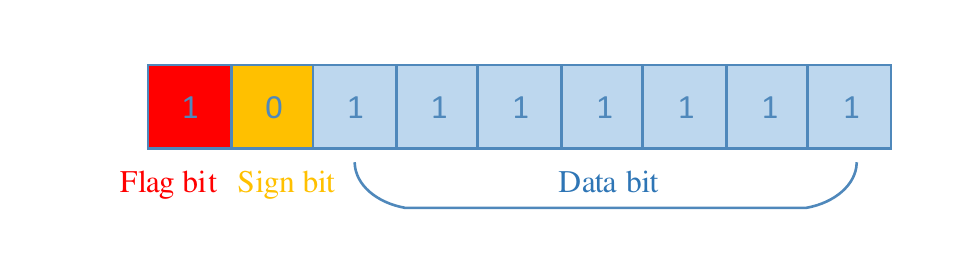}
        \label{fig:2b}}
    \caption {Data format of 9-bit integers.}
    \label{fig:3a}    
\end{figure}

By introducing a layer-wise scaling factor and a flag bit, we can expand the data coverage greatly. Details can be found in Figure \ref{fig:3a}. The flag bit is used to indicate whether the absolute value of $x$ stored in memory is less than the layer-wise scaling factor (e.g., 0 represents $|x| \textless Sc$ and 1 represents $|x| \geq Sc$). The sign bit is used to denote the positive or negative direction of the value. The data bit follows the conventional binary format. According to the definition, the values stored in Figure \ref{fig:2a} and \ref{fig:2b} are $+Sc/128$ and $-127 \times Sc$ when $k_{E_{2}}=8$, respectively. Therefore, the 9-bit data format can cover almost the same data range as the direct 15-bit quantization described in Equation (\ref{16bitE2}). Since the flag bit is just used for judgment, the effective value for computation is still INT8.

~\\
\indent \textit{(5) Gradient Quantization}
\vspace{4pt}

The gradient is another important part in DNN training because it is directly related to the weight update. The rules for calculating and quantizing the gradients of W, $\bm{\gamma}$, and $\bm{\beta}$ are described as Equation (\ref{G_define}) and (\ref{GQ_define}). Since $\bm{e_{1}}^{l}$, $\bm{e_{3}}^{l}$, $\bm{x_{0}}^{l}$, and $\bm{x_{2}}^{l}$ are all fixed-point values, the conventional FP MACs operations can be replaced with bit-wise operations during the process of calculating $\bm{g}_{W}^{l}$, $\bm{g}_{\bm{\gamma}}^{l}$, and $\bm{g}_{\bm{\beta}}^{l}$. The quantization functions are defined to further reduce the bit width of gradients and prepare for the next step of the optimizer. Specifically, we have
\begin{eqnarray}
\begin{split}
&Q_{G_{W}}(\bm{x})=CQ(\bm{x}, k_{G_{W}})\\
&~Q_{G_{\bm{\gamma}}}(\bm{x})=Q(\bm{x}, k_{G_{\bm{\gamma}}})\\
&~Q_{G_{\bm{\beta}}}(\bm{x})=Q(\bm{x}, k_{G_{\bm{\beta}}})\\
\end{split}
\label{QG_define}
\end{eqnarray}
where $CQ(\cdot)$ is the constant-quantization function defined in Equation (\ref{cqf}); $k_{G_{W}}$, $k_{G_{\bm{\gamma}}}$, and $k_{G_{\bm{\beta}}}$ are the bit width of the gradient of W, $\bm{\gamma}$, and $\bm{\beta}$, respectively.

~\\
\indent \textit{(6) Momentum Optimizer Quantization}
\vspace{4pt}

Momentum optimizer is one of the most common optimizers used in DNN training, especially for classification tasks. For the $i$-th training step of the $l$-th layer, the conventional Momentum optimizer works as follows
\begin{eqnarray}
\begin{split}
Acc_{i}^{l}=Mom \cdot Acc_{i-1}^{l}+\bm{g}_{i}^{l}
\end{split}
\label{Con_Mom}
\end{eqnarray}
where $Acc_{i}^{l}$ and $Acc_{i-1}^{l}$ are the accumulation in the $i$-th and $(i-1)$-th training step, respectively; $Mom$ is a constant value used as a coefficient; $\bm{g}_{i}^{l}$ is the gradient of W, $\bm{\gamma}$, or $\bm{\beta}$.

Momentum optimizer under the WAGEUBN framework is trying to constrain all operands to fixed-point values. The process can be formulated as 
\begin{eqnarray}
\begin{split}
&Acc_{i}^{l}=Mom \cdot Acc_{(i-1)q}^{l}+g^{l}_{iq} \\
&~~~~~~Acc_{iq}^{l}=Q_{Acc}(Acc_{i}^{l})
\end{split}
\label{Q_Mom}
\end{eqnarray}
where $Acc_{(i-1)q}^{l}$ is the quantized accumulation in the $(i-1)$-th training step; $g^{l}_{iq}$ is the quantized gradient of W, $\bm{\gamma}$, or $\bm{\beta}$; $Q_{Acc}(\cdot)$ is the quantization function defined as
\begin{equation}
Q_{Acc}(\bm{x})=Q(\bm{x}, k_{Acc}).
\label{QACC}
\end{equation}
To guarantee the consistency of bit width, we further set
\begin{equation}
 k_{G_{\bm{\gamma}}}=k_{G_{\bm{\beta}}}=k_{GC}=k_{Mom}+k_{Acc}-1.
 \label{GBit}
\end{equation}

~\\
\indent \textit{(7) Update Quantization}
\vspace{4pt}

The parameter update is the last step in the training of each mini-batch.  Different from conventional DNNs where the learning rate can take any FP value, the learning rate under WAGEUBN must also be a fixed-point value and the bit width of update is directly related to the bit width of learning rate. The update under quantized Momentum optimizer can be described as
\begin{eqnarray}
\begin{split}
&\Delta W = lr \cdot Acc_{i}^{l}\\
&W^{l} = W^{l} - \Delta W 
\end{split}
\label{Q_Update}
\end{eqnarray}
where $\Delta W$ is the update of W with $k_{WU}$ bits, and $lr$ is the fixed-point learning rate with $k_{lr}$ bits. The updates of $\bm{\gamma}$ and $\bm{\beta}$ are the same as in Equation (\ref{Q_Update}). According to Equation (\ref{Q_Mom}), (\ref{GBit}), and (\ref{Q_Update}), we have
\begin{eqnarray}
\begin{split}
&k_{WU}=k_{\bm{\gamma}U}=k_{\bm{\beta}U}=k_{Mom}+k_{Acc}+k_{lr}-2\\
&~~~~~~~~~~~~~~~~~~~~~~~=k_{GC}+k_{lr}-1\\
&~~~~~~~~~~~~~~~~~~~~~~~=k_{G_{\bm{\gamma}}}+k_{lr}-1\\
&~~~~~~~~~~~~~~~~~~~~~~~=k_{G_{\bm{\beta}}}+k_{lr}-1.
\end{split}
\label{UBit}
\end{eqnarray}
Through our evaluations, the precision of the update has the greatest impact on the accuracy of DNNs because it is the last step to constrain the parameters. Thus, we need to set a reasonable bit width for update to balance the model accuracy and memory cost.

\subsection{Quantization Framework}
\label{Quantization Framework}

\begin{figure*}[!htpb]
    \centering
    \includegraphics[width=0.99\textwidth]{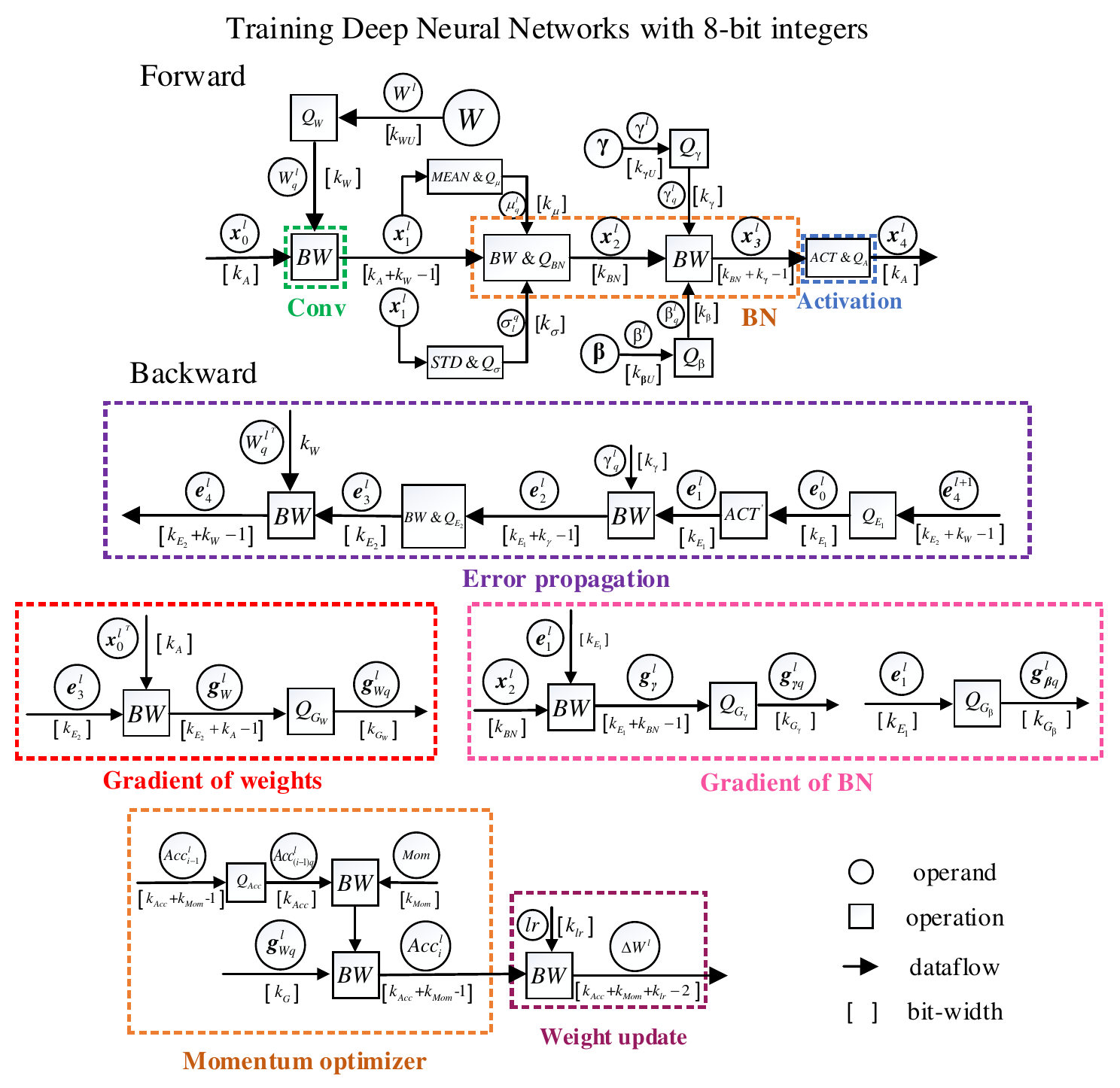}
    \caption {Overview of the WAGEUBN quantization framework. ``BW'' denotes bit-wise operations.}
    \label{fig:1}
\end{figure*}

Given the quantization details of W, A, G, E, U, BN, and the Momentum optimizer, the overall quantization framework is depicted in Figure \ref{fig:1}. Under this framework, conventional FP MACs can be replaced with bit-wise operations. Here, the forward pass of the $l$-th layer in DNNs is divided into three parts: Conv (convolution), BN, and activation. $\bm{x}_{0}^{l}$, $\bm{x}_{1}^{l}$, $\bm{x}_{2}^{l}$, $\bm{x}_{3}^{l}$, and $\bm{x}_{4}^{l}$ are defined in Equation (\ref{X_define}). The weights ($W^{l}$) are stored as $k_{WU}$-bit integers and then $Q_{W}$ maps $W^{l}$ to $k_{W}$-bit INT values ($W^{l}_{q}$) before convolution. After convolution, $MEAN\&Q_{\mu}$ and $STD\&Q_{\sigma}$ operations are used to calculate the mean and standard deviation of $\bm{x}_{1}^{l}$ in one mini-batch and then quantize them to $k_{\mu}$ and $k_{\sigma}$ bits, respectively. The $BW\&Q_{BN}$ operation constrains $\bm{x}_{2}^{l}$ to $k_{BN}$ bits in BN. Similar to W, $\bm{\gamma}$ and $\bm{\beta}$ are stored as $k_{\bm{\gamma}U}$ and $k_{\bm{\beta}U}$-bit integers and used in $k_{\bm{\gamma}}$ and $k_{\bm{\beta}}$ bits ($\bm{\gamma}^{l}_{q}$ and $\bm{\beta}^{l}_{q}$) after the $Q_{\bm{\gamma}}$ and $Q_{\bm{\beta}}$ quantization, respectively. After the second step of BN, activation and quantization are implemented with the $ACT\&Q_{A}$ operation, reducing the increased bit width to $k_{A}$ bits again and preparing inputs for the next layer.

The backward pass of the $l$-th layer is much more complicated than the forward pass, including error propagation, gradient of weight, gradient of BN, Momentum optimizer, and weight update. In the process of error propagation, $\bm{e}_{0}^{l}$, $\bm{e}_{1}^{l}$, $\bm{e}_{2}^{l}$, $\bm{e}_{3}^{l}$, and $\bm{e}_{4}^{l}$ are defined in Equation (\ref{E_define}) and there are two locations needing quantization using $Q_{E_{1}}$ and $Q_{E_{2}}$. $Q_{E_{1}}$ reduces the bit width of $\bm{e}_{4}^{l+1}$ from $k_{E_{2}}+k_{W}-1$ to $k_{E_{1}}$. $ACT'$ is the derivative of activation function ($relu$) and $Q_{E_{2}}$ is used to constrain $\bm{e}_{3}^{l}$ to $k_{E_{2}}$ bits. In the phase of calculating the gradients of weights and BN, $Q_{G_{W}}$, $Q_{G_{\bm{\gamma}}}$, and $Q_{G_{\bm{\beta}}}$ are leveraged to reduce the increased bit width caused by the multiplication operations.

All parameters of the Momentum optimizer in the $i$-th training step are quantized. Different from the conventional learning rate with FP value, WAGEUBN requires a discrete learning rate so that the bit width of weight updates can be controlled. The updates of $\bm{\gamma}$ and $\bm{\beta}$ in BN layers are also similar to the weight update, which are omitted in Figure \ref{fig:1} for simplicity.

\subsection{Overall Algorithm}
\label{Process of Algorithm}

    \begin{algorithm}[!htpb]
    \setstretch{1.35} 
    \caption{Forward pass of $l$-$th$ layer} 
    \label{alg:forward}
    \begin{algorithmic}
            \STATE    Convolution:
            \STATE ~~$\bm{x}_{0}^{l} \Leftarrow \bm{x}^{l-1}_{4}$, $\bm{W}_{q}^{l} \Leftarrow Q_{W}(\bm{W}^{l}, k_{W})$ 
            \STATE ~~$\bm{x}_{1}^{l} \Leftarrow \bm{W}_{q}^{l} \bm{x}_{0}^{l}$
            \STATE BN:
            \STATE ~~ $\mu^{l}_{q} \Leftarrow Q_{\mu}(\mu^{l}) $, $\sigma^{l}_{q} \Leftarrow Q_{\sigma}(\sigma^{l})$
            \STATE ~~ $\bm{x}_{2}^{l} \Leftarrow Q_{BN}(\dfrac{\bm{x}_{1}^{l}-\mu_{q}^{l}}{\sigma^{l}_{q}})$
            \STATE ~~ $\bm{\gamma}_{q}^{l} \Leftarrow Q_{\bm{\gamma}}(\bm{\gamma}^{l}) $, $ \bm{\beta}_{q}^{l} \Leftarrow Q_{\bm{\beta}}(\bm{\beta}^{l})$
            \STATE ~~ $\bm{x}_{3}^{l} \Leftarrow \bm{\gamma}_{q}^{l}\bm{x}_{2}^{l} + \bm{\beta}_{q}^{l} $
            \STATE Activation:
            \STATE ~~ $\bm{x}_{4}^{l} \Leftarrow Q_{A}(relu(\bm{x}_{3}^{l}))$
        \end{algorithmic} 
\end{algorithm}

    \begin{algorithm}[!htpb]
        \setstretch{1.5}
        \caption{Backward pass of $l$-$th$ layer} 
        \label{alg:backward}
        \begin{algorithmic}
            \STATE Error propagation:
            \STATE ~~ $\bm{e}_{0}^{l} \Leftarrow Q_{E_{1}}(\bm{e}_{4}^{l+1})$ 
            \STATE ~~ $\bm{e}_{1}^{l} \Leftarrow \bm{e}_{0}^{l} \odot \dfrac{\partial \bm{x}_{4}^{l}}{\partial \bm{x}_{3}^{l}}$        
            \STATE ~~ $\bm{e}_{2}^{l} \Leftarrow \bm{e}_{1}^{l} \odot \bm{\gamma}_{q}^{l}$        
            \STATE ~~ $\bm{e}_{3}^{l} \Leftarrow  Q_{E_{2}}(\bm{e}_{2}^{l} \odot \dfrac{\partial \bm{x}_{2}^{l}}{\partial \bm{x}_{1}^{l}})$        
            \STATE ~~ $\bm{e}_{4}^{l} \Leftarrow {W_{q}^{l}}^{T} \bm{e}_{3}^{l} $
            \STATE Gradient computation:
            \STATE ~~ $\bm{g}_{W}^{l} \Leftarrow \bm{e}_{3}^{l} {\bm{x}_{0}^{l}}^{T} $        
            \STATE ~~ $\bm{g}_{Wq}^{l} \Leftarrow Q_{G_{W}}(\bm{g}_{W}^{l}) $        
            
            \STATE ~~ $\bm{g}_{\bm{\gamma}}^{l} \Leftarrow \bm{e}_{1}^{l} \odot \bm{x}_{2}^{l} $        
            \STATE ~~ $\bm{g}_{\bm{\gamma}q}^{l} \Leftarrow Q_{G_{\bm{\gamma}}}(\bm{g}_{\gamma}^{l})$
            
            \STATE ~~ $\bm{g}_{\bm{\beta}}^{l} \Leftarrow \bm{e}_{1}^{l} $        
            \STATE ~~ $\bm{g}_{\bm{\beta}q}^{l} \Leftarrow Q_{G_{\bm{\beta}}}(\bm{g}_{\beta}^{l})$    
            
            
            \STATE Momentum optimizer($i$-$th$ step):
            \STATE~~ $Acc_{i}^{l}=Mom \cdot Acc_{(i-1)q}^{l}+\bm{g}^{l}_{iq}(\bm{g}_{Wq}^{l}, \bm{g}_{\bm{\gamma}q}^{l}, or ~\bm{g}_{\bm{\beta}q}^{l})$
            \STATE ~~ $Acc_{iq}^{l} = Q_{Acc}(Acc_{i}^{l})$
            
            \STATE Weight updates($i$-$th$ step):

            \STATE ~~ $\Delta W = lr \cdot Acc_{i}^{l}$
            \STATE ~~ $W^{l} = W^{l} - \Delta W$
            
        \end{algorithmic}
    \end{algorithm}

Given the framework of WAGEUBN, we summarize the entire quantization process and present the pseudo codes for both the forward and backward passes as shown in Algorithm \ref{alg:forward} and \ref{alg:backward}, respectively.

\section{Results}
\label{Sec:Exp}

\subsection{Experimental Setup}

To verify the effectiveness of the proposed quantization framework, we apply WAGEUBN on ResNet18/34/50 on ImageNet dataset. We provide two versions of WAGEUBN, one with full 8-bit INT where $k_{W}$, $k_{A}$, $k_{G_{W}}$, $k_{E_{1}}$, $k_{E_{2}}$, $k_{\bm{\gamma}}$, and $k_{\bm{\beta}}$ are equal to 8.
The other version has 16-bit $k_{E_{2}}$. The only difference between the 16-bit $E_{2}$ version and the full 8-bit version exists in the quantization function $Q_{E_2}$ (see Equation (\ref{16bitE2}) and (\ref{8bitE2}), respectively). $k_{G_{\bm{\gamma}}}$, $k_{G_{\bm{\beta}}}$, and $k_{GC}$ are 15 and we set $k_{mom}$, $k_{Acc}$, $k_{lr}$, and $k_{WU}$ to 3, 13, 10, and 24 respectively to satisfy Equation (\ref{GBit}) and (\ref{UBit}). In addition, we set $k_{BN}$, $k_{\mu}$, and $k_{\sigma}$ to 16. Since W, A, G, and E occupy the majority of memory and compute costs, their bit-width values are reduced as much as possible. Other parameters occupying much less resources can increase the bit width to maintain the accuracy, e.g., $\mu_{q}^{l}$, $\sigma_{q}^{l}$, $g_{\bm{\gamma q}}^{l}$, and $g_{\bm{\beta q}}^{l}$ in BN layers.

The first and last layers are believed to differ from the rest because of their interface with network inputs and outputs. The quantization of these two layers will cause significant accuracy degradation compared to hidden layers and they just consume few overheads due to the small number of neurons. Therefore, we do not quantize the first and last layers, as previous work did \cite{Wan2018,choi2018}.

For the quantization frameworks, it's a common issue of saturated learning and vanishing gradients because the data in DNNs may be clipped very small after quantization and the model may be stuck on a local optimum because of the few updates. And this situation can be more serious when the quantization framework is used in large-scale complex datasets. WAGEUBN has made many efforts to alleviate this problem. Firstly, the newly designed shift-quantization function ensures that the quantized error in WAGEUBN framework is on the same magnitude as that of a traditional FP network, making the errors (gradients of activations) not vanishing in the process of back propagation. Secondly, based on the observation that it is the orientation rather than the magnitude of gradients (gradients of weights) that guides deep neural networks (DNNs) to converge, it's more important to keep the orientation rather than the magnitude. What's more, the updates ($\Delta W$) jointly determined by the learning rate and gradients are the data that ultimately determines the final changes of the network. The much bigger learning rate setting in WAGEUBN (minimum learning rate: WAGEUBN  $1.95\times 10^{-3}$ ($2^{-9}$) VS floating-point network $5\times10^{-6}$ ) can make up for the problem of update magnitude caused by the smaller quantized gradients. Through the technics above, the problems of saturated learning and dead gradients can be solved. In addition to the above measures which can help the model get rid of local optimum, the proposed quantized Momentum optimizer, fixed-point updates which is precise enough and a proper batch size selection can also reduce the probability of falling into a local optimum greatly.

\subsection{Training Curve}

\begin{figure}[htpb]
    \centering
    \includegraphics[width=0.46\textwidth]{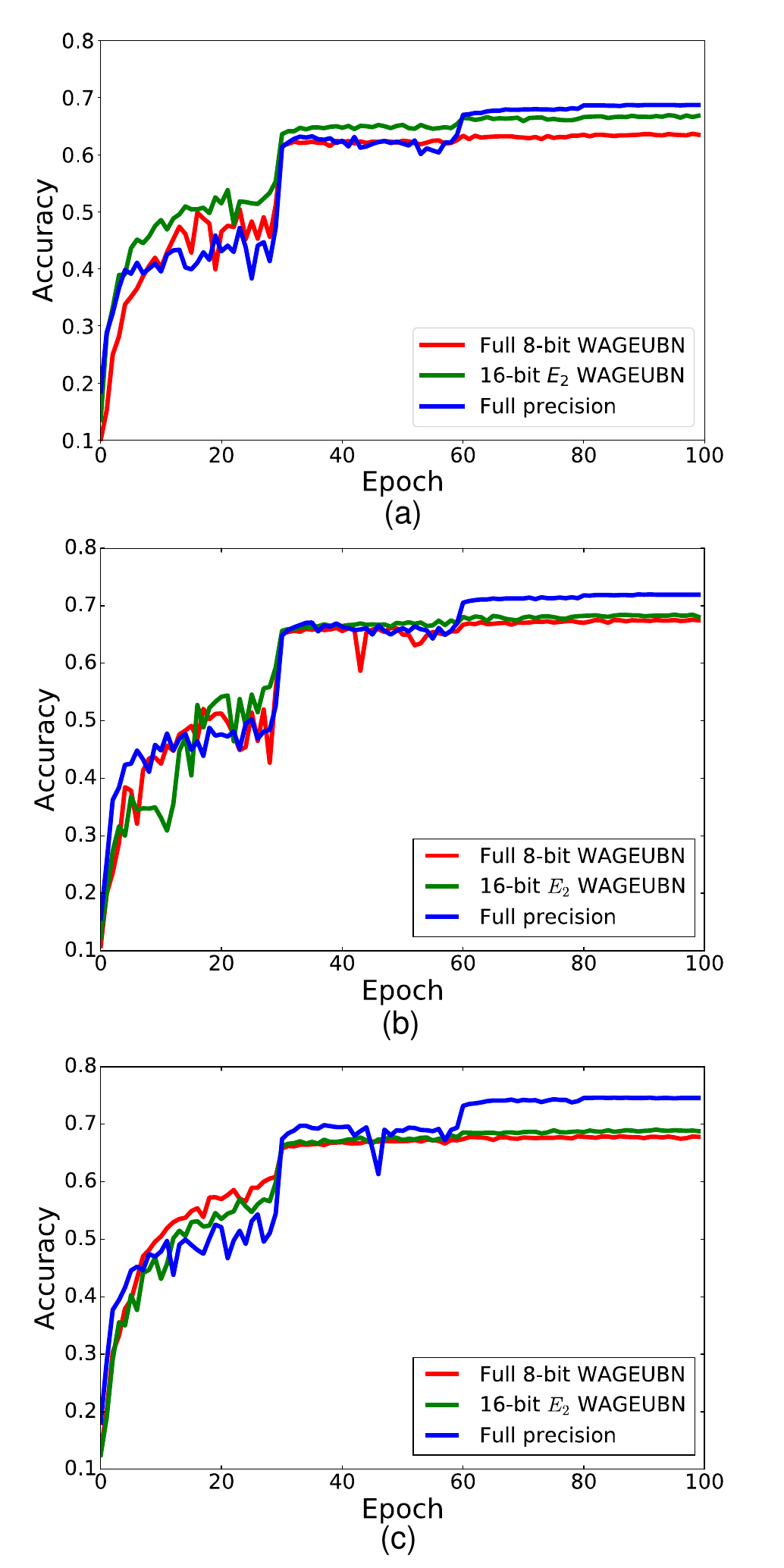}
    \caption {Training curves under the WAGEUBN framework: (a) ResNet18; (b) ResNet34; (c) ResNet50.}
  \label{Fe8}
\end{figure}

Figure \ref{Fe8} illustrates the accuracy comparison between vanilla DNNs (FP32), DNNs with the full 8-bit version of WAGEUBN, and DNNs with the 16-bit $E_{2}$ version of WAGEUBN. 
The initial learning rate and momentum coefficient under WAGEUBN and full precision are slightly different. To ensure the best recognition accuracy, we use the official parameter settings of TensorFlow \cite{abadi2016tensorflow}, where the initial learning rate and momentum coefficient are set to 0.05 (batch size is 128) and 0.9 while those of WAGEUBN framework are set to 0.05078125 ($26\times2^{-9}$, 10-bit integer) and 0.75 ($3\times2^{-2}$, 3-bit integer). And this may cause the phenomenon that WAGEUBN converges faster at the beginning of training. During epoch 30 and 60, we have reduced the learning rate, which is a general practice in training process \cite{brownlee2016using,lau2017learning}.
The training curves show that there is little difference between vanilla DNNs and the ones under the WAGEUBN framework when the training epoch is less than 60, which reflects the effectiveness of our approach. As the epoch evolves, the accuracy gap begins to grow because the learning rate in vanilla DNNs is much lower than that in WAGEUBN, such as $5\times 10^{-6}$ v.s. $1.95\times 10^{-3}$ ($k_{lr}=10$), thus the update of vanilla DNNs is more precise than that under the WAGEUBN framework. We can further improve the accuracy by reducing the learning rate, while the bit width values of learning rate and update need to increase accordingly at the expense of more overheads.

\begin{table*}
    \caption{Accuracy of vanilla DNNs and WAGEUBN DNNs on ImageNet dataset.}
    \label{Acc}
    \vspace{-10pt}
    \begin{center}
    \begin{spacing}{1.2}
        \begin{tabular}{c|cccccccc}
            \hline
            Network& $k_{W}$& $k_{A}$& $k_{G_{W}}$& $k_{E_1}$& $k_{E_2}$& $k_{WU}$& Accuracy Top-1/Top-5(\%)\\
            \hline
            & 32& 32& 32& 32& 32& 32& 68.70/88.37 \\
            ResNet18& 8& 8& 8& 8& 16& 24& 67.40/87.63 \\
            & 8& 8& 8& 8& 8& 24& 64.79/85.81 \\
            \hline
            & 32& 32& 32& 32& 32& 32& 71.99/90.56 \\
            ResNet34& 8& 8& 8& 8& 16& 24& 68.50/87.96 \\
            & 8& 8& 8& 8& 8& 24& 67.63/87.70 \\
            \hline
            & 32& 32& 32& 32& 32& 32& 74.66/92.13 \\
            ResNet50& 8& 8& 8& 8& 16& 24& 69.07/88.45 \\
            & 8& 8& 8& 8& 8& 24& 67.95/88.01 \\
            \hline    
        \end{tabular}
    \end{spacing}
    \end{center}
\end{table*}

Table \ref{Acc} quantitatively presents the accuracy comparison between vanilla DNNs and WAGEUBN DNNs on the ImageNet dataset. We have achieved the state-of-the-art accuracy on large-scale DNNs with full 8-bit INT quantization. The 16-bit $E_2$ WAGEUBN only loses 3.46\% mean accuracy compared with the vanilla DNNs. Because the bit width of most data keeps the same between the full 8-bit WAGEUBN and the 16-bit $E_2$ WGAEUBN, the overhead difference between them is negligible. The DNNs under WAGEUBN framework have achieved an accuracy that is comparable to FP8 \cite{Wang2018} and QBP2 \cite{Banner2018}. And although MP \cite{Micikevicius2017} and MP-INT \cite{Das2018} can achieve the accuracy close to the full precision networks, the computational cost is much higher than WAGEUBN because of the floating-point data type and higher bit width, which will be detailed in Section \ref{Sec:Dis}. Compared with the vanilla DNNs, about $4 \times$ memory size shrink, much faster processing speed, and much less energy and circuit area can be achieved under the proposed WAGEUBN framework.

\subsection{Quantization Strategies for W, A, G, E, and BN}

\begin{figure*}[!htpb]
    \centering
    \includegraphics[width=0.99\textwidth]{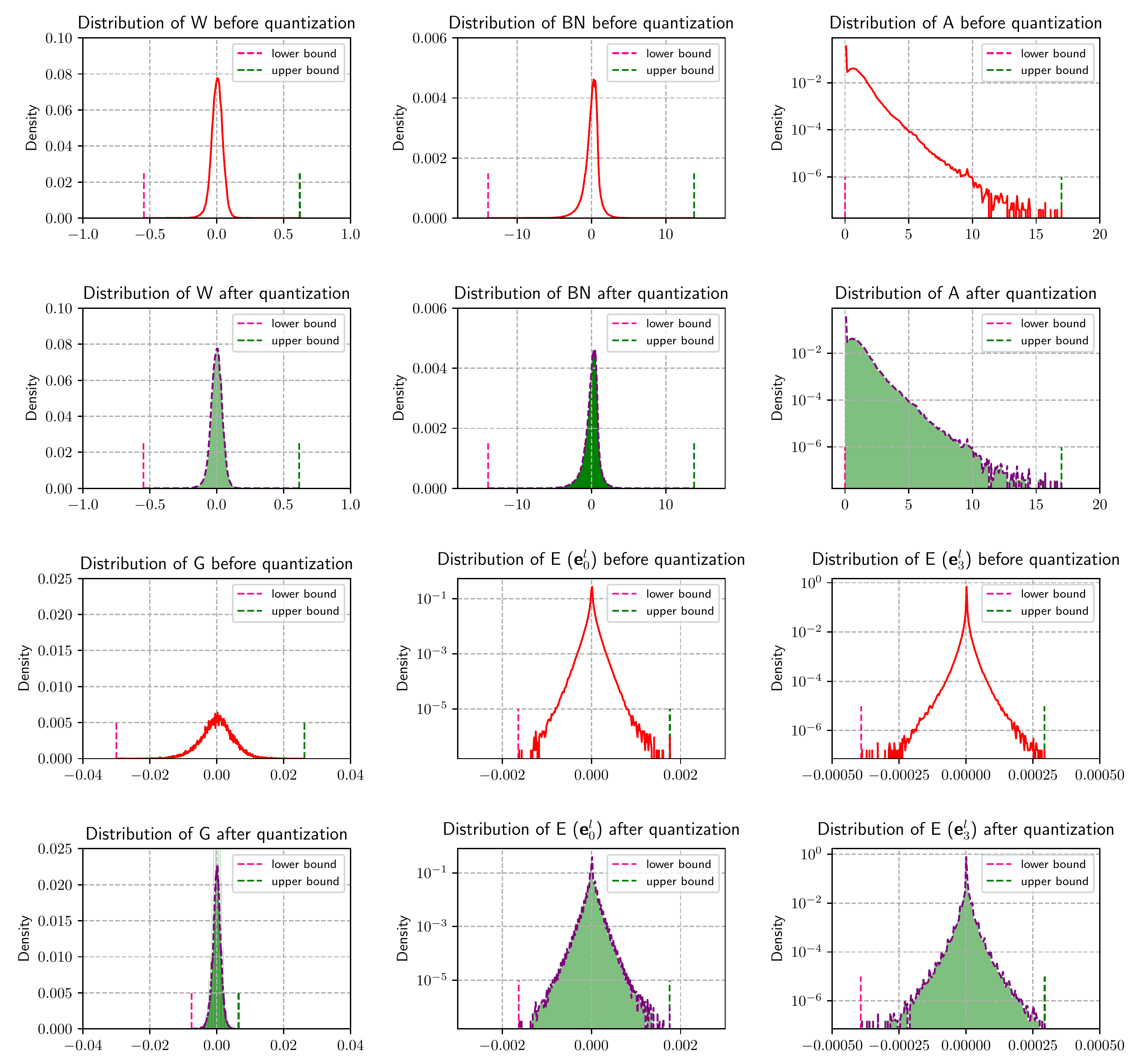}
    \caption {Data distribution comparison between W, BN, A, G, and E  before and after quantization.}
    \label{fig:WAGEBN}
\end{figure*}

In our WAGEUBN framework, we use different quantization strategies for W, A, G, E, and BN, i.e. $Q(\cdot)$ for W, A, and BN; $CQ(\cdot)$ for G, and $SQ(\cdot)$ for E. Different quantization strategies are based on the data distribution, data sensitivity, and hardware friendliness. Figure \ref{fig:WAGEBN} shows the distribution comparison between W, BN ($\bm{x}_{2}^{l}$ defined in Equation (\ref{X_define})), A, G (weight gradient), and E ($\bm{e}_{0}^{l}$, $\bm{e}_{3}^{l}$ defined in Equation (\ref{E_define})) before and after quantization.

According to the definition, the resolution of the direct-quantization function is $2^{-7}$ when the bit width equals 8 and there is no limitation on the data range. Because W, BN, and A in the inference stage directly affect the loss function and further influence the backpropagation, the quantization of W, BN, and A should be as precise as possible to avoid the loss fluctuation. This is guaranteed for the reason that the resolution of the direct-quantization function is enough for W, BN, and A, which indicates that the direct-quantization function barely changes their data distributions.

The constant-quantization function has a resolution of $2^{-14}$ and the data range after quantization is about $[-2^{-7}, 2^{-7}]$ in the case of $k_{GC}=15$ and $k=8$. $k$ will decrease as the training epoch goes on, causing the data range reduction. Figure \ref{fig:WAGEBN} reveals that the constant-quantization function changes the data distribution of G greatly while the network accuracy has not declined much as a result. The reason behind this phenomenon is that it is the orientation rather than the magnitude of gradients that guides DNNs to converge. In the meantime, it is easy to ensure that the bit width of updates can be fixed when $k_{GC}$ is fixed, which is more hardware-friendly since the bit width of weights stored in memory can also be fixed during training. 

The shift-quantization retains the magnitude order and omits the general values whose absolute value is less than $2^{-7}$ when $k=8$. The 8-bit shift-quantization works well for the quantization of error after activation ($\bm{e}_{1}^{l}$ defined in Equation (\ref{E_define})). However, we find the shift-quantization is not enough for the quantization of errors between Conv and BN ($\bm{e}_{3}^{l}$ defined in Equation (\ref{E_define})). Therefore, the newly designed quantization function in Equation (\ref{8bitE2}) named 8-bit Flag $Q_{E_2}$ is utilized. Figure \ref{fig:WAGEBN} shows that the distribution of E ($\bm{e}_{3}^{l}$) is almost the same before and after quantization, revealing the validity of the 8-bit Flag $Q_{E_2}$ quantization function.

\subsection{Accuracy Sensitivity Analysis} 

To compare the influences of W, A, G, E, and BN quantization individually, we quantize them to 8-bit INT separately with the FP32 update. Taking $k_{W}=8$ as an example, we quantize only W to 8-bit INT and leaving others (A, BN, G, E, and U) still kept in FP32. The quantization function for single data used here is the same as what Section \ref{Quantization details of WAGEUBN} describes (Equation (\ref{8bitE2}) is used for the error quantization when $k_{E_{2}}=8$).

\begin{table*}[!htpb]
    \caption{Accuracy sensitivity under WAGEUBN with single data quantization on ResNet18.}
    \label{8bWAGEBN}
    \vspace{-10pt}
    \begin{center}
        \begin{tabular}{c|c|c|c|c|c|c}
            \hline
            Bit-width& $k_{W}=8$& $k_{BN}=8$& $k_{A}=8$& $k_{G_{W}}=8$& $k_{E_{1}}=8$& $k_{E_{2}}=8$ \\
            \hline
          Accuracy Top-1/Top-5 (\%)& 67.98/88.02& 68.01/87.96& 67.74/87.89& 67.88/87.89& 67.88/87.92& 67.08/87.44\\
            \hline    
        \end{tabular}
    \end{center}
\end{table*}

The results of ResNet18 under the WAGEUBN framework with single data quantization is shown in Table \ref{8bWAGEBN}. The accuracy of single data quantization reflects the difficulty degree when quantizing W, A, G, E, and BN, separately. From the table, we can see that the quantization of E, especially $\bm{e}_{3}^{l}$ defined in Equation (\ref{E_define}), makes the most impacts on accuracy. In addition, we find that the accuracy heavily fluctuates during training when $\bm{e}_{3}^{l}$ is constrained to 8-bit INT, which does not appear in the quantization of other data. 
To sum up, the E data, especially $\bm{e}_{3}^{l}$, demands the highest precision and is the most sensitive component under our WAGEUBN framework.

\begin{figure}[!htpb]
    \centering
    \includegraphics[width=0.46\textwidth]{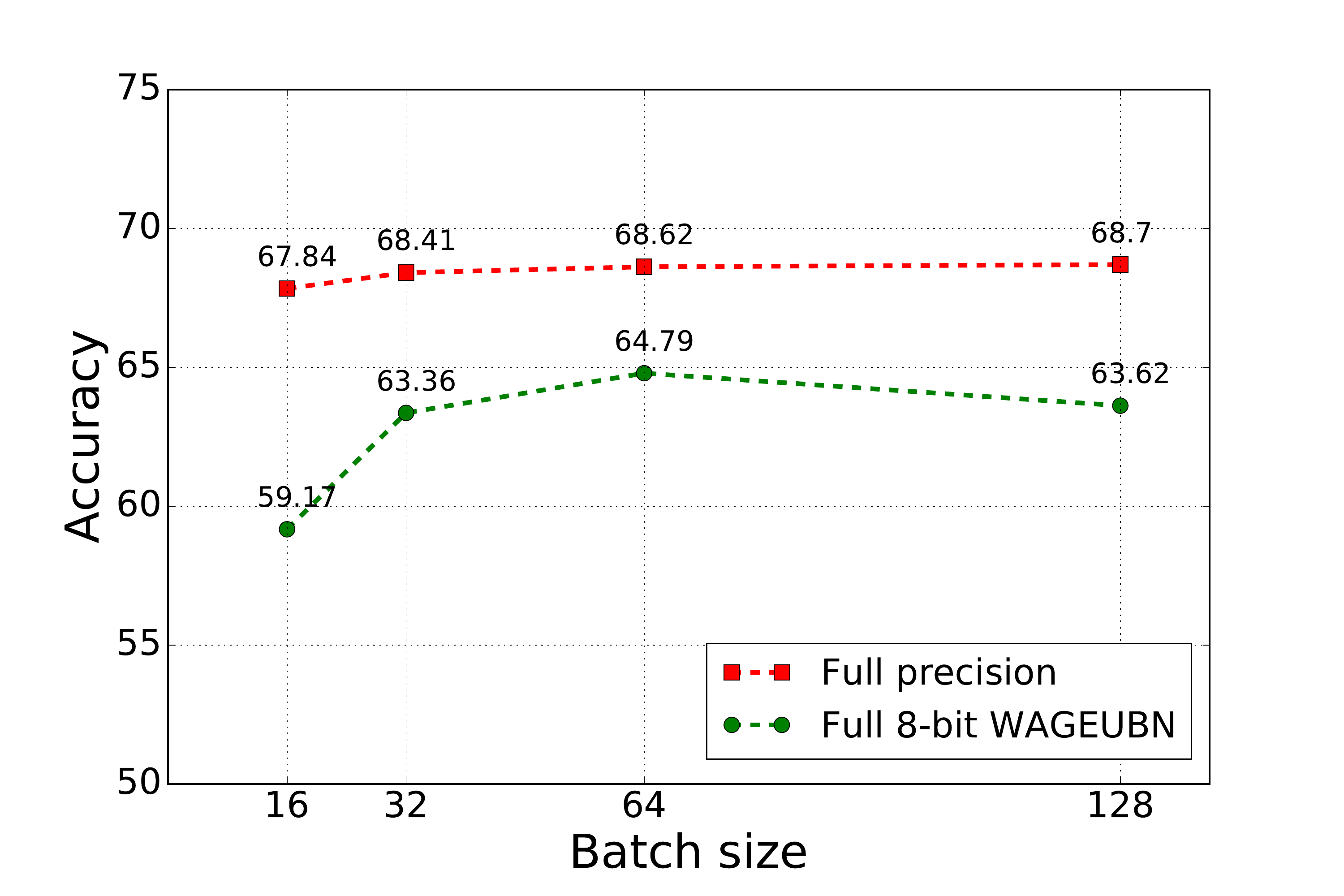}
    \caption {Accuracy sensibility analysis of batch size under WAGEUBN.}
    \label{fig:bs}
\end{figure}

Because the BN process in the forward pass and the average gradients in the backward pass all involve the batch size, the sensibility of batch size to network performance under WAGEUBN is also explored. We have done comparative experiments and the results are illustrated in Figure \ref{fig:bs}. When the batch size changes between 128 and 16, the accuracy of the full precision DNNs won't drop significantly, while that of the full 8-bit WAGEUBN DNNs has a relatively large decline in accuracy when the batch size equals 16. One reason for this lies in that there is a momentum used to update the moving means and standard deviations in a batch for full precision DNNs while WAGEUBN abandons this considering the computational cost. Besides, we think it is normal because quantization will inevitably result in the loss of some precise information. And the loss of information will bring a slightly higher batch size sensitivity. However, the robustness of WAGEUBN is still quite good, because there is a significant decline in accuracy only when the batch size is less than a very small number 16.

\subsection{Analysis of the Error Quantization between Conv and BN}

Error backpropagation is the foundation of DNN training. If the error of E caused by quantization is too large, the convergence of DNNs will be degraded.
Especially, because the error quantization between convolution and BN ($\bm{e}_{3}^{l}$) is directly related to the weight update of the $l$-th layer, the impact of $\bm{e}_{3}^{l}$ quantization on the model accuracy is critical. 

\begin{figure}[!htpb]
    \centering
    \includegraphics[width=0.5\textwidth]{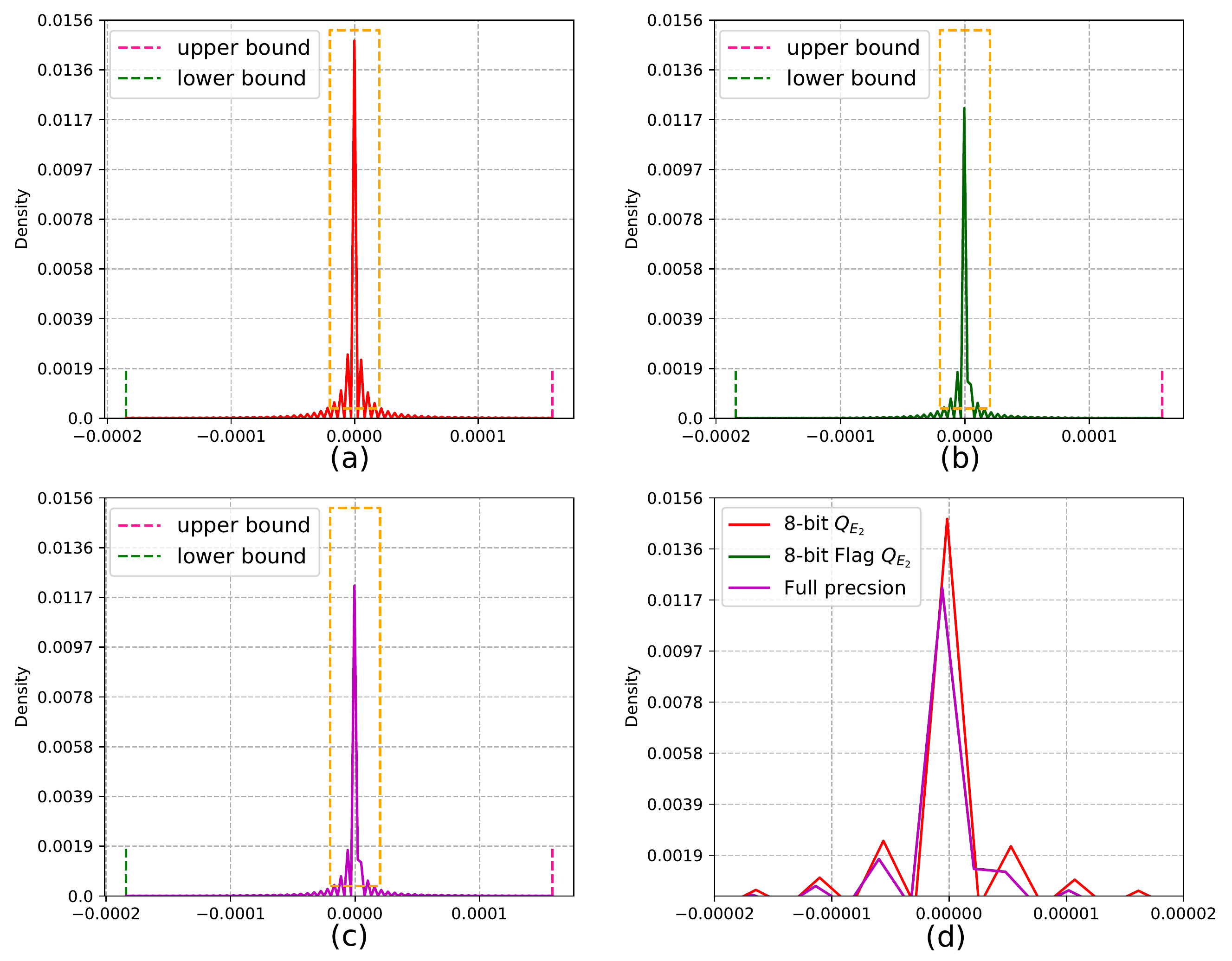}
    \caption {Data distributions of $\bm{e}_{3}^{1}$: (a) 8-bit $Q_{E_{2}}$, (b) 8-bit Flag $Q_{E_{2}}$, (c) full precision; (d) Data distribution comparison.}
    \label{fig:E89}
\end{figure}

To further analyze the reason why 8-bit $Q_{E_{2}}$ (defined in Equation (\ref{16bitE2}) where $k_{E_{2}}=8$) causes the non-convergence of DNNs and compare the distributions of $\bm{e}_{3}^{l}$ under 8-bit $Q_{E_{2}}$, 8-bit Flag $Q_{E_{2}}$ (defined in Equation (\ref{8bitE2}) where $k_{E_{2}}=8$), and full precision, the data distribution of $\bm{e}_{3}^{l}$ of the first quantized layer on ResNet18 is shown in Figure \ref{fig:E89}. From the figure, we can see that the $\bm{e}_{3}^{1}$ distributions under 8-bit $Q_{E_{2}}$ quantization and full precision differs a lot and those of 8-bit Flag $Q_{E_{2}}$ quantization and full precision are almost the same. The major difference between the 8-bit $Q_{E_{2}}$ quantization and full precision lies in the interval of $[-{2^{-8}}R(\bm{e}_{3}^{1}),{2^{-8}}R (\bm{e}_{3}^{1})]$ ($R(\cdot)$ is defined in Equation (\ref{cqf})), where the 8-bit $Q_{E_{2}}$ quantization forces the data in this range to zero.

\label{BWE}

\begin{figure}[!htpb]
    \centering
    \includegraphics[width=0.5\textwidth]{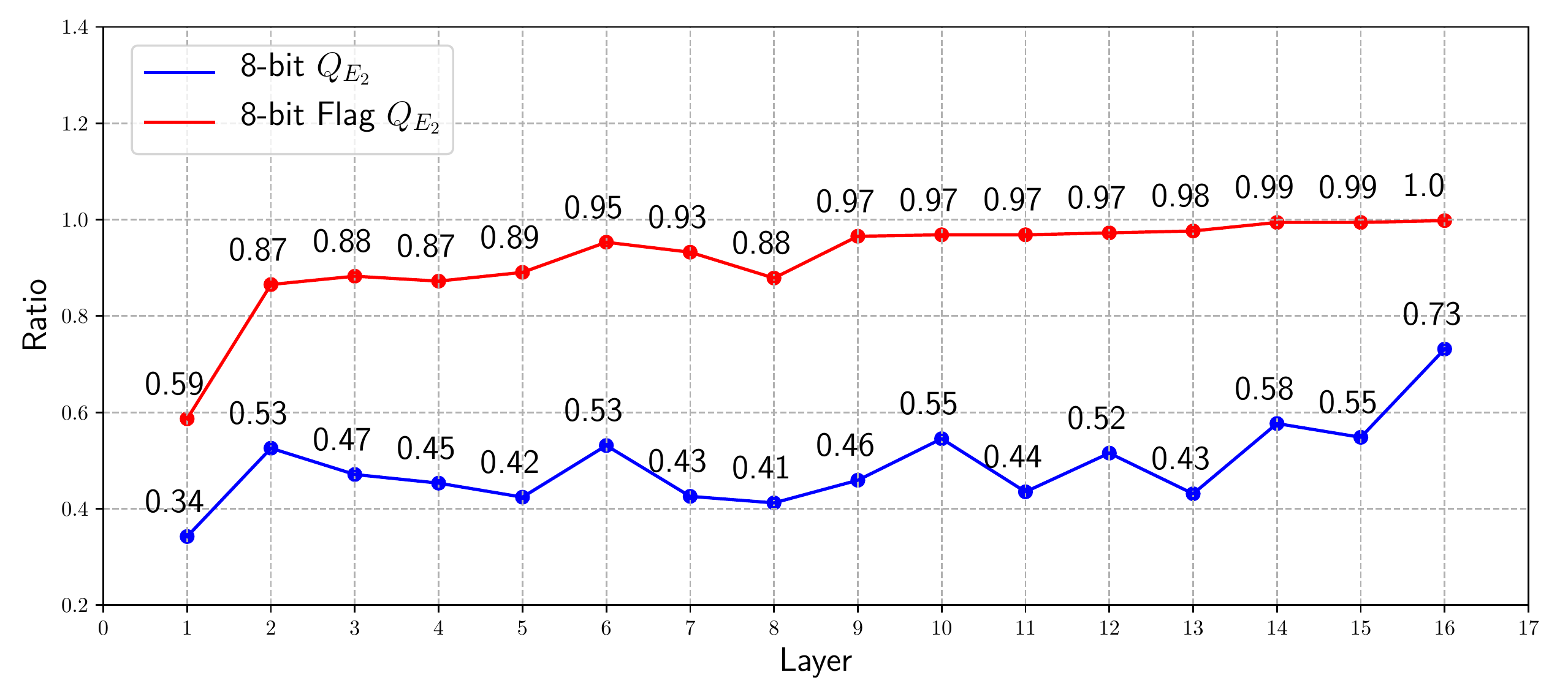}
    \caption {Data ratios of 8-bit $Q_{E_{2}}$ and 8-bit Flag $Q_{E_{2}}$ quantization methods covered by each layer of ResNet18.}
    \label{fig:R89}
\end{figure}

The only difference between 8-bit $Q_{E_{2}}$ and 8-bit Flag $Q_{E_{2}}$ quantization functions lies in the data range. Theoretically, the covered data range of 8-bit $Q_{E_{2}}$ and 8-bit Flag $Q_{E_{2}}$ are about $[-R(\bm{e}_{3}^{l}), -{2^{-8}}R(\bm{e}_{3}^{l})]\cup\{0\}\cup[{2^{-8}}R(\bm{e}_{3}^{l}), R(\bm{e}_{3}^{l})]$ and $[-R(\bm{e}_{3}^{l}), -{2^{-15}}R(\bm{e}_{3}^{l})]$ $\cup$ $\{0\}$ $\cup$ $[{2^{-15}}R(\bm{e}_{3}^{l}), R(\bm{e}_{3}^{l})]$, respectively. Because the distribution of $\bm{e}_{3}^{l}$ is not uniform, the range covered by different quantization methods varies a lot. The data ratios (Proportion of non-zero values after quantization.) of 8-bit $Q_{E_{2}}$ and 8-bit Flag $Q_{E_{2}}$ quantization functions covered by each layer of ResNet18 are illustrated as Figure \ref{fig:R89}. 
Although the larger values take greater impacts on the model accuracy in the process of error propagation, the smaller values also contain useful information and occupy the majority. Compared with 8-bit Flag $Q_{E_{2}}$, the data ratio 8-bit $Q_{E_{2}}$ covers is too little because of the smaller data range. That is to say, although the most important information contained by the larger values is retained, the information contained by the smaller values is ignored, resulting in the non-convergence of DNNs. In addition, there is also a rough trend that the data ratio decreases as the network becomes shallower, either in the 8-bit $Q_{E_{2}}$ or 8-bit Flag $Q_{E_{2}}$ quantization.

\subsection{Cost Discussion}
\label{Sec:Dis}

\begin{figure}[!htpb]
    \centering
    \subfigure[]{\includegraphics[width=0.48\textwidth]{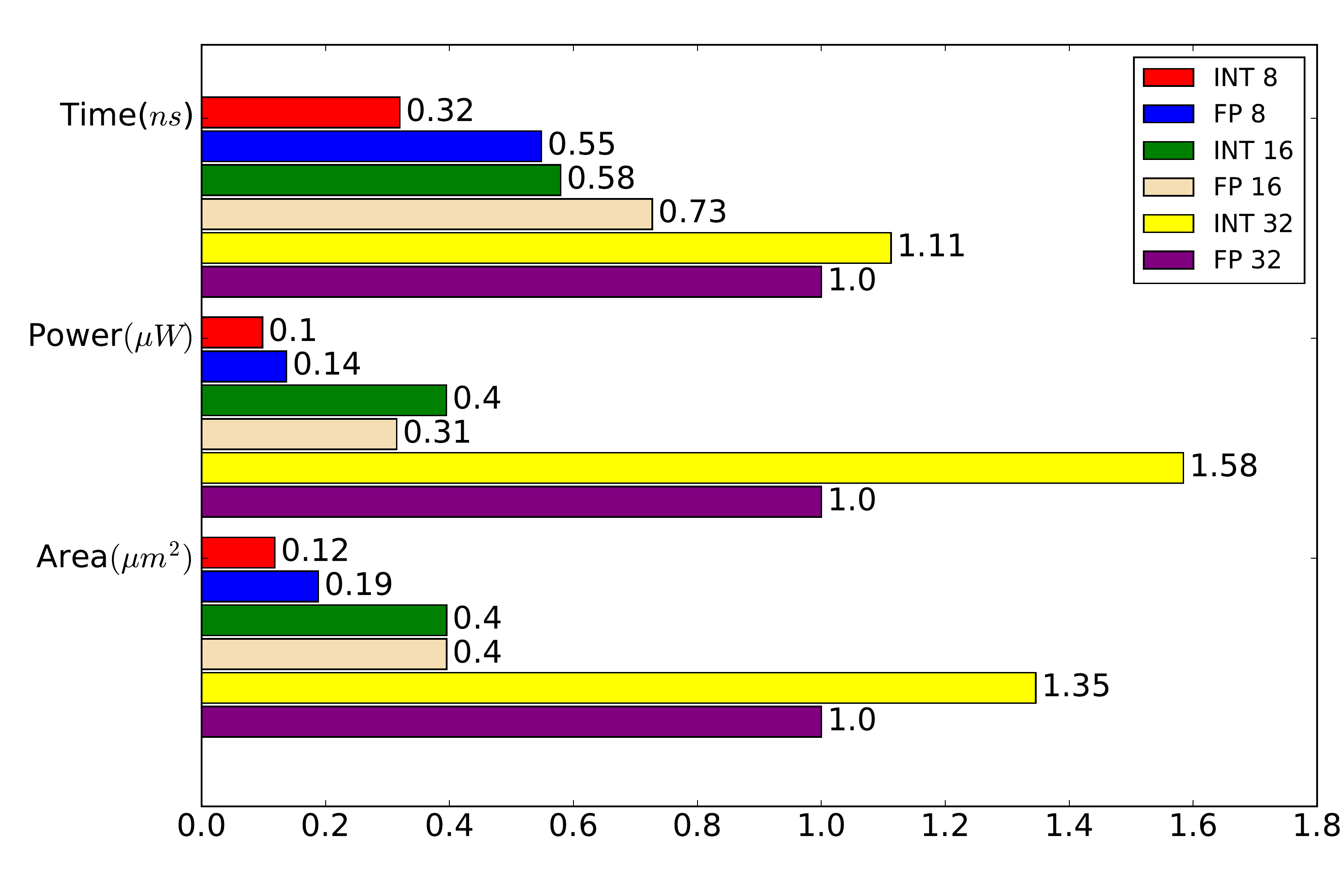}
        \label{fig:SPAM}}
    \subfigure[]{\includegraphics[width=0.48\textwidth]{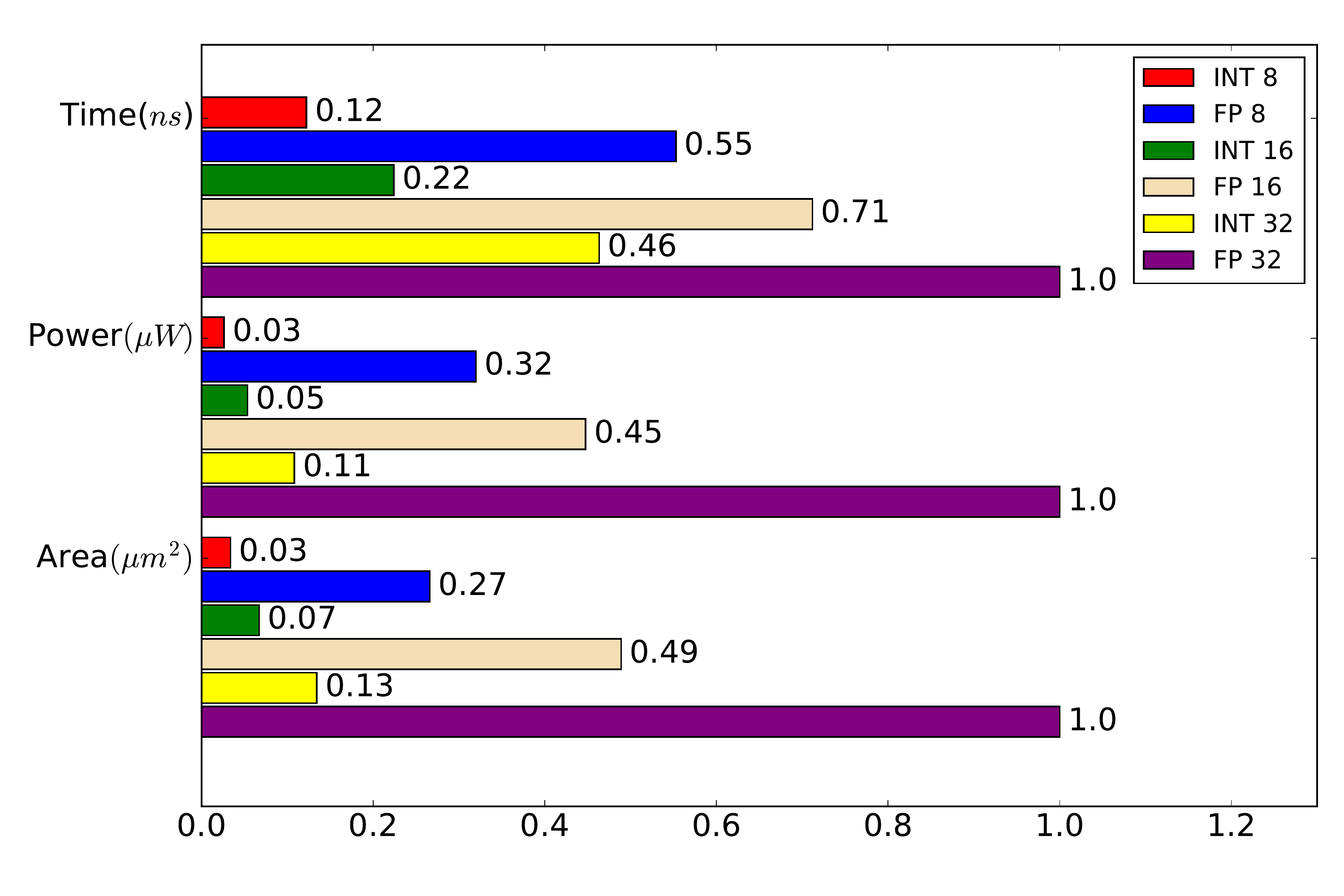}
        \label{fig:SPAA}}
    \caption {Comparison of time, power, and area of single multiplication and accumulation operation under different quantization precision: (a) multiplication, (b) accumulation.}
    \label{fig:SPA}
\end{figure}

Although it is recognized that DNN quantization can greatly reduce memory and compute costs, resulting in lower energy consumption, quantitative analysis is rarely seen in recent research. In order to compare the full INT8 quantization with other precision solutions (FP32, INT32, FP16, INT16, and FP8) more clearly, we have simulated the processing speed, power consumption, and circuit area for single multiplication and accumulation operation on FPGA platform. Figure \ref{fig:SPA} shows the results. With FP32 as the baseline, taking the multiplication operation as an example, INT8 can perform  $>$3$\times$ faster in speed, 10$\times$ lower in power, and 9$\times$ smaller in circuit area. Similarly, compared with FP32, the speed of INT8 accumulation is about 9$\times$ faster, and the energy consumption and circuit area are reduced by $>$30$\times$. In addition, the INT8 multiplication and accumulation operations are more advantageous than other data type operations, whether it is FP8\cite{Wang2018}, INT16\cite{Das2018}, FP16\cite{Micikevicius2017} or INT32. In conclusion, the proposed full INT8 quantization has great advantages in hardware overheads, whether in terms of memory cost, processing speed, power consumption, and circuit area. Given the huge advantages in hardware resources and computing speed of quantization, the idea of low-precision computing and the quantization functions in WAGEUBN can also extend to other research fields which involve large-scale matrix operations, such as large-scale control systems \cite{tharanidharan2019finite}, weather forecasting models \cite{vavna2017single}, etc.

\section{CONCLUSIONS}
\label{Sec:Con}

We propose a unified framework termed as ``WAGEUBN'' to achieve a complete quantization of large-scale DNNs in both training and inference with competitive accuracy. We are the first to quantize DNNs over all data paths and promote DNN quantization to the full INT8 level. In this way, all the operations can be replaced with bit-wise operations, causing significant improvements in memory overhead, processing speed, circuit area, and energy consumption. Extensive experiments evidence the effectiveness and efficiency of WAGEUBN. This work provides a feasible solution for the online training acceleration of large-scale and high-performance DNNs and further shows the great potential for the applications in future efficient portable devices with online learning ability. Although great progress has been made, WAGEUBN still needs to build a special hardware architecture design and develop more applications. Future works could transfer to the design of computing architecture, memory hierarchy, interconnection infrastructure, and mapping tool to enable the specialized machine learning chips and more applications.

\section*{ACKNOWLEDGMENT}

The work was supported partially by the National Natural Science Foundation of China (No.
61603209, 61876215).

\bibliographystyle{unsrt}
\bibliography{library}

\end{document}